\documentclass[journal]{IEEEtran}
\IEEEoverridecommandlockouts                              
\usepackage[utf8]{inputenc}
\usepackage[T1]{fontenc}
\usepackage{graphicx}
\usepackage[namelimits]{amsmath} 
\usepackage{amssymb}             
\usepackage{amsfonts}            
\usepackage{mathrsfs}            
\usepackage{subfigure} 
\usepackage{booktabs}
\usepackage{cite}
\usepackage{url}
\usepackage{dsfont}

\usepackage{algorithm}
\usepackage{algpseudocode}
\usepackage{algorithmicx}
\usepackage{amsmath}
\usepackage[switch]{lineno}

\usepackage{multirow}
\usepackage{hyperref}
\usepackage{amsthm}
\newtheorem{theorem}{Theorem}

\newtheorem{lemma}{Lemma}

\usepackage{textcomp,booktabs}
\usepackage[usenames,dvipsnames]{color}
\usepackage{colortbl}
\definecolor{mgray}{gray}{.9}

\usepackage[colorinlistoftodos,bordercolor=blue,backgroundcolor=blue!20,linecolor=blue,textsize=scriptsize]{todonotes}
\usepackage{xcolor, soul}
\sethlcolor{yellow}

\newcommand{\Ch}[1]{\textcolor{Maroon}{\textbf{#1}}}
\newcommand{\change}[1]{\textcolor{RoyalBlue}{\textbf{#1}}}

\title{Augmenting Reinforcement Learning with Transformer-based Scene Representation Learning for Decision-making of Autonomous Driving}

\author{Haochen Liu, Zhiyu Huang, Xiaoyu Mo, and Chen Lv$^{*}$,~\IEEEmembership{Senior Member, IEEE} 
\thanks{H. Liu, Z. Huang, X. Mo, and C. Lv are with the School of Mechanical and Aerospace Engineering, Nanyang Technological University, 639798, Singapore. (E-mails: {\tt\small haochen002@e.ntu.edu.sg}, {\tt\small zhiyu001@e.ntu.edu.sg}, {\tt\small xiaoyu006@e.ntu.edu.sg}, {\tt\small lyuchen@ntu.edu.sg})}
\thanks{This work was supported by the SUG-NAP Grant (No. M4082268.050) of Nanyang Technological University, Singapore.}
\thanks{$^{*}$Corresponding author: C. Lv}
}

\begin{document}
\maketitle
\thispagestyle{empty}
\pagestyle{empty}

\begin{abstract} Decision-making for urban autonomous driving is challenging due to the stochastic nature of interactive traffic participants and the complexity of road structures. Although reinforcement learning (RL)-based decision-making schemes are promising to handle urban driving scenarios, they suffer from low sample efficiency and poor adaptability. In this paper, we propose the Scene-Rep Transformer to enhance RL decision-making capabilities through improved scene representation encoding and sequential predictive latent distillation. Specifically, a multi-stage Transformer (MST) encoder is constructed to model not only the interaction awareness between the ego vehicle and its neighbors but also intention awareness between the agents and their candidate routes. A sequential latent Transformer (SLT) with self-supervised learning objectives is employed to distill future predictive information into the latent scene representation, in order to reduce the exploration space and speed up training. The final decision-making module based on soft actor-critic (SAC) takes as input the refined latent scene representation from the Scene-Rep Transformer and generates decisions. The framework is validated in five challenging simulated urban scenarios with dense traffic, and its performance is manifested quantitatively by substantial improvements in data efficiency and performance in terms of success rate, safety, and efficiency. Qualitative results reveal that our framework is able to extract the intentions of neighbor agents, enabling better decision-making and more diversified driving behaviors. Code and results are available at: \change{{\href{https://georgeliu233.github.io/Scene-Rep-Transformer/}{https://georgeliu233.github.io/Scene-Rep-Transformer/}}}
\end{abstract}


\section{Introduction}
\label{sec1}
\IEEEPARstart{M}{aking} safe, smooth, and intelligent decisions is a major challenge for autonomous vehicles (AVs) \cite{huang2021driving, b2}, especially in complex urban driving scenarios \cite{b3}.  The sophisticated interaction dynamics among various traffic participants and road structures make it difficult to achieve these goals. Reinforcement learning (RL)-based methods have gained popularity in addressing these challenges \cite{b4, huang2022efficient, wu2021prioritized}. RL agents can learn from interactions with driving environments (mostly in simulation) and are capable of handling diverse traffic situations and tasks, freeing researchers and engineers from burdensome rule-based decision schemes \cite{b5,b6}. However, RL-based approaches still encounter some common issues that may impede their further development. One leading drawback is data efficiency, which means that training a functional driving policy requires a massive amount of samples through interactions with the environment. This problem is exacerbated when facing urban driving scenarios, such as unprotected left turns or road branches with dense traffic. Another problem is how to construct and encode the state representation of urban driving scenarios, which is a bottleneck in generalizing RL to more complex scenes. Unlike common RL problems where the state of the environment or agent is relatively simple, state representations for urban driving should cover heterogeneous information such as dynamic features among traffic flows, latent agent interactions, and road structures. Moreover, distilling predictive information in the scene representation can help the agent understand the consequences of certain decisions, facilitating the learning of driving policies for decision-making.


Accordingly, this paper proposes a Transformer-based structure, named Scene-Rep Transformer, to better capture the relationships among heterogeneous elements and inject predictive information into the scene representation, in order to enhance the ability of the RL-based decision-making schemes and expedite training. First, we construct a multi-stage Transformer (MST) that encodes raw heterogeneous scene element information, such as the agent's historical motion states and candidate route waypoints, and fuses the information to represent their interactions. The MST produces a single latent feature vector about the driving scene at a given time step. Second, we introduce the sequential latent Transformer (SLT) to distill the common features from a sequence of latent-action pairs from future time steps into the latent at the current step, using a self-supervised learning method. The purpose of SLT is to acquire a core understanding of the sequential dynamics of the environment and reduce the exploration space because the current latent representation contains consistent information about future steps. The SLT is only used in the training process to facilitate the learning of latent vectors and speed up training. Eventually, the SAC-based decision-making module takes as input the refined latent feature vector and outputs better-quality decisions. The main contributions of this paper can be outlined as follows:
\begin{enumerate}
\item A novel Transformer-based scene representation learning framework is designed to effectively augment the ability of RL-based decision-making systems in urban driving scenarios, both in training and testing.
\item A multi-stage Transformer is proposed to encode the information of heterogeneous scene elements and model the interactions among agents and map waypoints, resulting in a latent scene representation for the RL module. A sequential latent Transformer with an augmented self-supervised learning objective is proposed to distill the predictive sequential information into the latent representation to facilitate RL training.
\item {The proposed framework is analytically proved with performance boosting.} It is further validated under multiple challenging interactive driving scenarios and shows a significant effect in improving the RL driving agent's performance in terms of data efficiency, task success rates, and interpretability. The effects of different components of our framework are also investigated.
\end{enumerate}

\section{Related Work}
\subsection{Reinforcement Learning for Autonomous Driving}
Reinforcement learning (RL) recently has been widely employed in decision-making tasks for autonomous driving to good effect \cite{b4}, outcompeting the rule-based approaches that are usually onerous and cannot function well in complex tasks \cite{b2}. Another competitive approach is imitation learning \cite{huang2020multi, chen2019deep}, which relies on supervised learning with expert data, but it suffers from the distributional shift problem in the deployment. On the other hand, RL can avoid this problem because it relies on interacting with the environment but may require a massive amount of interactions to train an acceptable policy. Therefore, recent works have proposed some methods to improve the data efficiency of RL, such as guiding the RL decision-making with expert actions \cite{wu2021human,huang2022efficient,b12} as it can significantly reduce the action exploration space. In our work, rather than using expensive expert guidance or demonstration, we propose to leverage the representation learning technique to obtain a better, predictive, and generalizable representation of the environment, in order to improve the training data efficiency and testing performance of RL. Another problem is the fitly selection of the action space for driving decision-making. One mainstream makes end-to-end decisions \cite{b6,perez2022deep}, of which map a continuous action space directly with steering \cite{wolf2017learning}, throttle, or endpoints\cite{feher2019hybrid}; Whilst it is stuck on generality with discontinuities. \cite{b5}. Therefore, in our work, we alleviate the action settings by introducing strategic decision-making \cite{duan2020hierarchical} in various autonomous driving scenarios, which shows promising results.

Another solution for RL is sequential learning. Expert sequences of state-action-reward pairs are propagated in auto-regressive manners using Transformer decoders. TT \cite{janner2021reinforcement} infers discretized sequence mapping for each step using beam search, while DT \cite{chen2021decision} learns the action mapping. Comparable to IL, those methods are hungry for expert data, and go under reconstructing futures with verbosity. Therefore, we turn to distill reconstruction-free predictive latent as an auxiliary task for RL using SLT.


\subsection{Representation Learning in Reinforcement Learning}
A good scene representation is essential to aid the agent in comprehending the intricacies of its environment and improving its decision-making abilities. An intuitive approach was to map a policy with raw sensory data directly using an end-to-end network \cite{bojarski2016end}. However, uncontrollable results led to the consensus for decomposition into perception, decision-making, and control modules \cite{b6}. End-to-end methods now test their performance on hyper-realistic simulators \cite{dosovitskiy2017carla} by jointly learning all modules \cite{casas2021mp3}. Focusing on decision-making tasks, we in turn utilize intermediate-level observations after the perception process \cite{b5}. The driving scene with multi-modal information (e.g., map and traffic agents) inputs is initially represented in a rasterization format (images), which has been used in visual-based driving systems and shows simplicity and adaption. However, rasterized images inevitably lose the explicit relations between the agents and map \cite{b14}, which may cost a large network and more data to recover certain relations. A parallel technical route is the unified vectorized representation for both traffic agents and road maps \cite{b13}, which is widely applied in frontier motion forecasting task \cite{b15, gilles2022gohome}. In our work, we employ the dense vectorized representation that unifies the positions of agents, and waypoints of HD maps \cite{bender2014lanelets} under online graph search. It dynamically provides the local intentions of each agent that correlate map information to the decision-making tasks. Another important point for scene representation is modeling the interactions between different elements (e.g., agent-agent and agent-map). Some works have explored graph neural networks (GNNs) in the RL framework to formulate the interactions between multiple agents. For example, DeepScene-Q \cite{b18-2} introduced GNN-based encoder modeling interactions for urban driving decision-making. DQ-GAT further improved the DeepScene-Q method by introducing the graph attention network (GAT) \cite{b18-3}. In this work, we employ a more powerful Transformer-based structure for interaction modeling, which has been widely used for encoding the environment in the motion forecasting field \cite{b17, gilles2022gohome, b15}.

\begin{figure*}[ht]
    \centering
    \includegraphics[width=\linewidth]{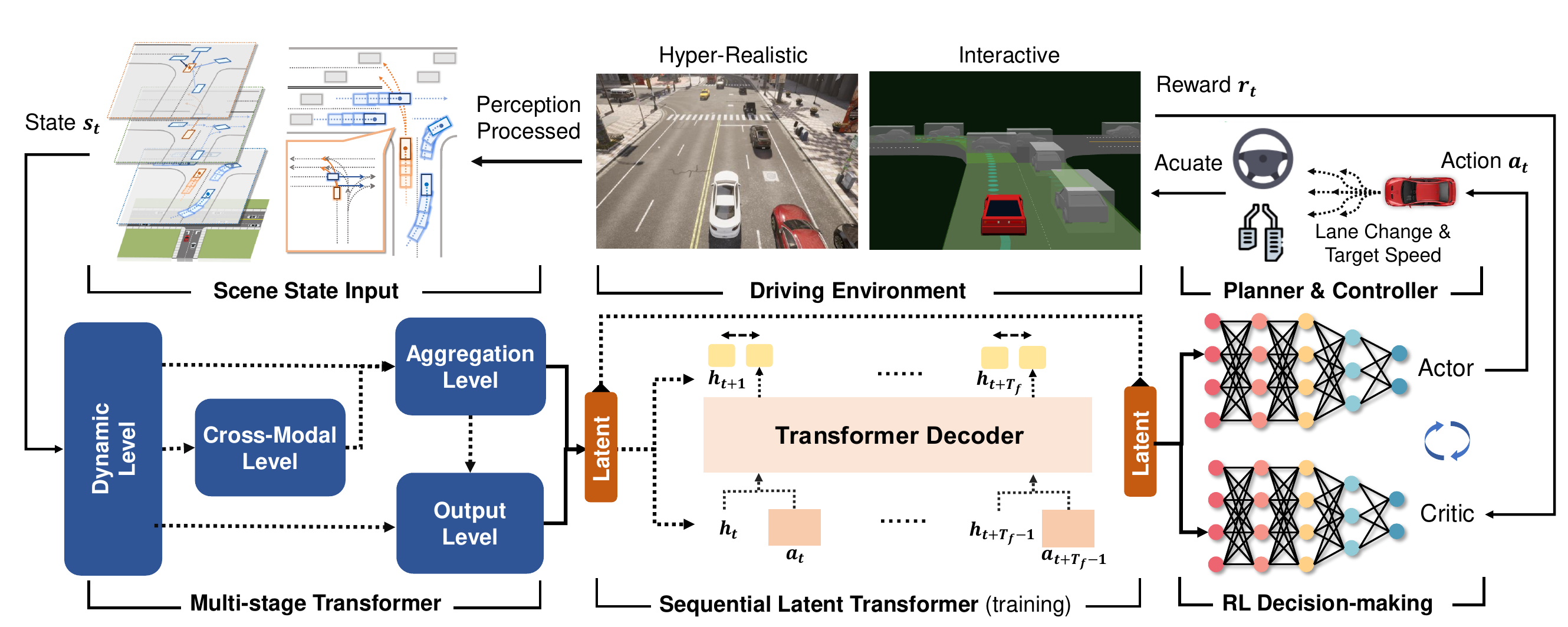}
    \caption{An overview of our RL decision-making framework with Scene-Rep Transformer. Given perception-processed vectorized scene inputs, the multi-stage Transformer (MST) encodes the multi-modal information with interaction awareness. During training, a sequential latent Transformer (SLT) performs representation learning using consecutive latent-action pairs to ensure future consistency. The soft-actor-critic (SAC) module takes the latent feature vector to make driving decisions for downstream planning and control tasks. Experiments are conducted in both hyper-realistic and interactive driving environments.}
    \vspace{-0.5cm}
    \label{fig:fig.1}
\end{figure*}

In addition to more effective environment encoding, another aspect to elevate the performance of RL through representation learning is to distill the predictive knowledge of environmental dynamics into the latent representation. Global value function (GVF) provides a predictive value for scene representations \cite{b23}, but may face importance sampling issues and have limited improvement facing sparse reward in complex urban scenes. \cite{b27} proposed a model-based imitative method to learn the global predictive knowledge from latent space. Further, an augmentation method for learning self-consistency \cite{b28} latent representation is proposed in DrQ \cite{b29} and CURL \cite{b30} and has been applied in image-based driving systems. Self-latent consistency is initially acquired through autoencoder-based approaches \cite{b25} by introducing the reconstruction loss, which imposes a great amount of computation burden on the system. In contrast, the self-supervised representation learning method eliminates the reconstruction loss with directly self-latent state pairing through a Siamese encoder \cite{b26}. However, most of the self-supervised approaches only consider the current or uni-step consistency to the prediction of global dynamics; yet the sequential consistency is another key factor for better representations under the partially observable environment. Therefore, in this work, we construct an augmentation method for representation learning and follow the idea of self-predictive representations (SPR) \cite{b31}, which achieves excellent performance in Atari games. Rather than a recurrent structure for global prediction, we adopt a Transformer to better adapt to the partially observable environment for autonomous driving.

\section{Reinforcement Learning with Scene-Rep Transformers}
\subsection{Framework}
\label{overview}
The proposed RL framework focuses on smart decision-making given a complex state input representing the dynamics and map information of the autonomous driving agent and its neighbors. This section provides an overview of the proposed framework, which consists of three main parts, as shown in Fig. \ref{fig:fig.1}. The multi-stage Transformer takes as input the vectorized scene representation, encodes the information and interactions among agents, and outputs a latent feature vector. The sequential latent Transformer then takes consecutive latent feature vectors across time together with corresponding action embedding and outputs a sequence of predictive latent vectors, which are trained with sequential consistency. The soft-actor-critic-based decision-making module finally takes the latent vector from the scene encoder and generates decision-making outputs. The trained multi-stage transformer and decision module are used during the inference (deployment) phase, while the sequential latent Transformer is employed for representation learning during training \cite{b28}.

\textbf{Problem Formulation:} Two objectives are considered during the learning process. For RL decision-making, the objective follows the Markov decision process (MDP) represented as a tuple $(s_t,a_t,r_t,s_{t+1},\gamma)$. Specifically, the self-driving agent receives state representation $s_{t}\in\mathcal{S}$ from the perception of the environment and executes an action $a_{t}\in\mathcal{A}$ according to its policy $\pi(a_{t}|s_t)$. The environment then returns a reward $r_t(s_t, a_t)$ and transitions to the next state $s_{t+1}\in\mathcal{S}$. The goal of RL is to optimize the policy $\pi$ so as to maximize the expected cumulative discounted returns: $ \max \ \mathbb{E}_{\pi} \left[ \sum \gamma^{t} r_t\right]$. The objective of self-supervised learning in the sequential latent Transformer module is to enforce the sequential similarities between the predicted latent $h_t^\prime$ and the future ground-truth one $h_{t+1}$ over a global predictive horizon $T_G$: $\min \ \frac{1}{T_G} \sum_t^{T_G} \mathcal{L}_{sim} ({h^\prime}_t, h_{t+1})$. 

\begin{figure*}[ht]
    \centering
    \includegraphics[width=\linewidth]{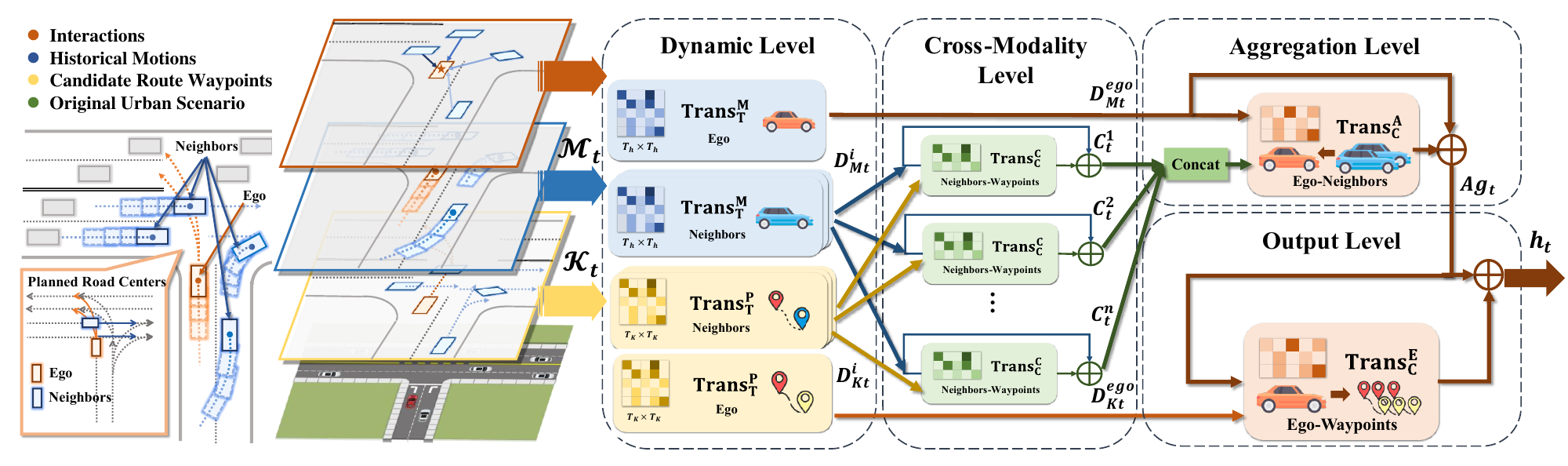}
    \vspace{-0.5cm}
    \caption{An overview of the MST Structure. The urban scenario on the left side can be layered to explicit representations $s_t$ as the historical motion states $\mathcal{M}_t$ and candidate route waypoints $\mathcal{K}_t$. The implicit layer for agent interactions is modeled in the subsequent multi-stage structure: \textbf{1) Dynamic level} encodes $\mathcal{M}_t$ and $\mathcal{K}_t$ along temporal axis; \textbf{2) Cross-modality level} models the interactions of neighbor agents, which are aggregated to the ego agent through \textbf{3) Aggregation level}; \textbf{4) Output level} adds the ego-waypoints and ego-neighbor interaction features to output the latent representation $h_t$.}
    \vspace{-0.5cm}
    \label{fig:fig.2}
\end{figure*}

\textbf{State representations:}  At a time step $t$, the state input $s_t$ encompasses the historical motion states of all agents and their local candidate route waypoints: $$s_t = [\mathcal{M}_t; \mathcal{K}_t],$$ where $\mathcal{M}_t=\{M^{ego}_{t}, M^{1}_{t}, M^{2}_{t}, \cdots, M^{n}_{t}\}$ contains historical motion trajectories of the controlled ego vehicle $M^{ego}_{t}$ and nearby $n$ vehicles $M^{1:n}_{t}$ within a certain distance $d_n$. Each motion trajectory is a sequence of motion states of certain vehicle over historical horizon $T_h$: $M_t=\{m_{t-T_h+1}, m_{t-T_h+2}, \cdots, m_{t}\}$, and each motion state $m_{t}$ includes positions, velocities, and heading angles: $m_t=(x_t, y_t, {v_x}_t, {v_y}_t, \psi_t)$. $\mathcal{K}_t$ represents candidate routes (See Section \ref{appendix}), $\mathcal{K}_t = \{K^{ego}_{t}, K^{1}_{t}, K^{2}_{t}, \cdots, K^{n}_{t}\}$ corresponding to different agents in the scene. The candidate route waypoints $K^{ego}_t, K^{n}_t$ for each agent consists of a set of local candidate route sequences ahead of its current position $K^{n}_t = \{\textbf{k}^{n,1}_t, \textbf{k}^{n,2}_t, \cdots\}$. Each local candidate route is composed of a sequence of waypoints across the future horizon $T_K$: $\textbf{k}_t=\{k_t, \cdots, k_{t+T_K}\}$, where $k_t=({w_x}_t, {w_y}_t, {w_\psi}_t)$ is the position and heading angle of a waypoint. It is a succinct representation of an agent's intention since it can outline the agent's future positions through the attention mechanism and improve the interpretability of decision-making. The state representation $s_t$ is encoded by the proposed multi-stage Transformer, followed by an off-policy RL algorithm for decision-making.

This work is focused on decision-making rather than designing an end-to-end autonomous driving system. As such, the perception and control modules will not be discussed. Instead, the state representations from the perception modules are taken as inputs, and our framework outputs for downstream control.

\subsection{Multi-stage Transformer}
\label{LPT}
The architecture of the multi-stage Transformer (MST) functions as a hierarchical encoder to map the scene states $s_t$ to a latent representation: $h_t\leftarrow \operatorname{\Phi_{MST}}(s_t)$. The structure is designed following the intuitions that 1) the model should be interaction-aware to handle interactions among agents in the driving scene; 2) the model should consider different modalities of interactions. Therefore, a Transformer-based network is employed to model the interactions among the map and agents in the driving scene, and a multi-stage structure is implemented for explicitly encoding the different modalities of interactions. The proposed MST structure for state encoding benefits the learning of latent representations in several ways: 1) the underlying dynamics of the driving scene are better captured; 2) the candidate route information is effectively leveraged for the guidance of future intentions; 3) the computation efficiency is guaranteed due to the quadratic complexity of the Transformer network; 4) the attention mechanism can perfectly deal with the varying number of agents in the scene, with excellency in interpretability. 

We will first introduce the multi-head attention, which is the core mechanism in the proposed MST structure for representing the interactions inside the Transformer at different levels. 
\begin{equation}
\label{e0}
\operatorname{MHA}(\mathbf{Q}, \mathbf{K}, \mathbf{V}, \mathbf{M}) = \text{concat} (\operatorname{head}_{1}, \cdots, \text{head}_{h})W^{O},
\end{equation}
where the attention results from $h$ heads are concatenated to capture different aspects of information. Each head captures the attention weighted value given $\mathbf{Q},\mathbf{K},\mathbf{V}$ denoting query, key, and value transformed by matrices $W_{i}^{Q}, W_{i}^{K}, W_{i}^{V}$:
\begin{equation}
\label{e1}
\text {head}_{i}=\operatorname{Att}(\mathbf{Q} W_{i}^{Q}, \mathbf{K} W_{i}^{K}, \mathbf{V} W_{i}^{V},\mathbf{M}).
\end{equation}
\begin{figure}[htp]
    \centering
    \includegraphics[width=\linewidth]{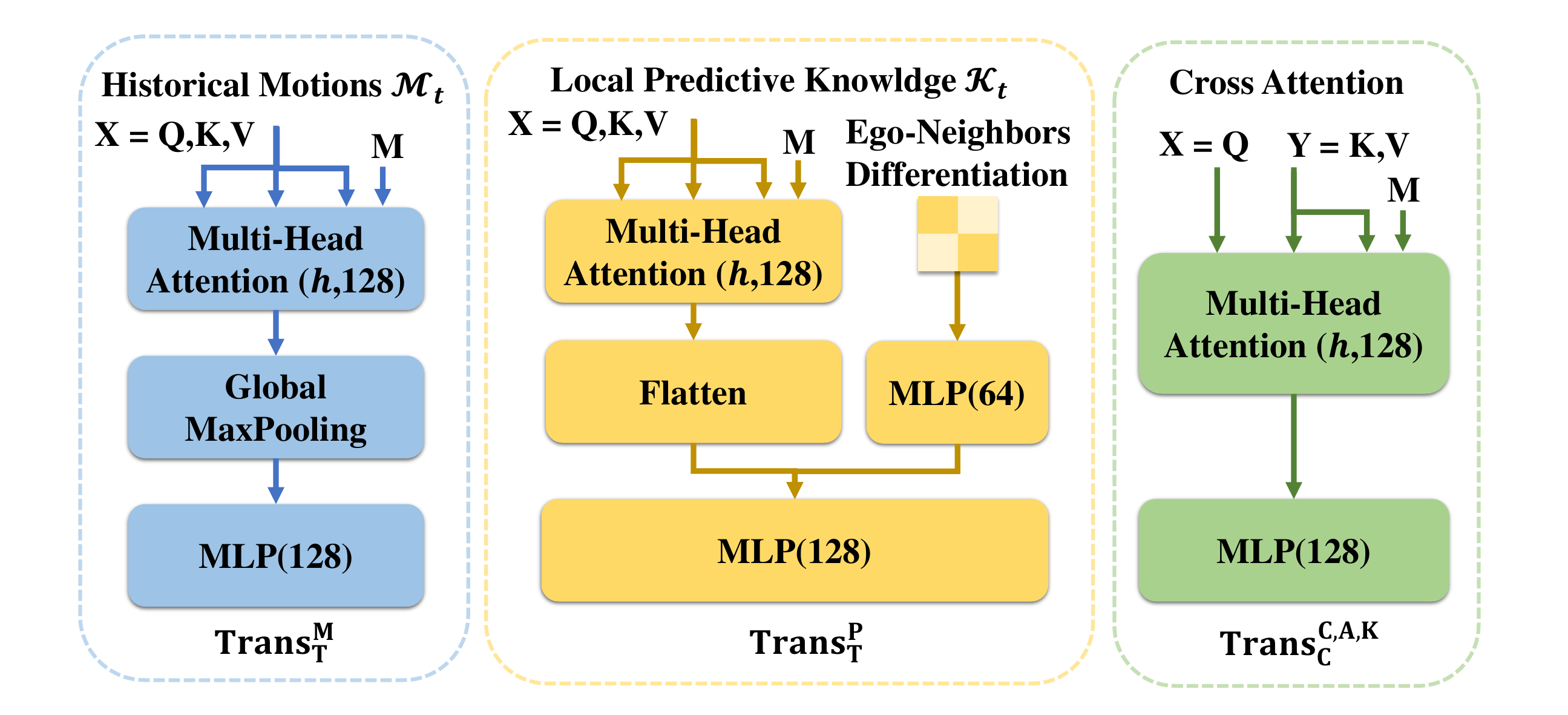}
    \caption{The structures of networks along different axes in our proposed MST.}    
    \label{fig:fig.3}
\end{figure}
 The attention-weighted values are calculated by scaled dot-product with attention mask $\mathbf{M}$:
\begin{equation}
\label{e2}
\operatorname{Att}(\mathbf{Q},\mathbf{K},\mathbf{V},\mathbf{M})=\operatorname{softmax}\left(\frac{\mathbf{Q K}^{T}}{\sqrt{d_{k}}}\cdot\mathbf{M}\right) \mathbf{V},
\end{equation}
where $d_{k}$ is the dimension of key. Attention mask $\mathbf{M}\in d_q\times d_k$ enables varying numbers of agents, and thus the highly dynamic nature of urban scenes is well-adapted across different elements and time steps.

Next, we will provide the detailed MST multi-stage structure, as illustrated in Fig. \ref{fig:fig.2}. The framework follows the subsequent computational flow: 1) dynamic level that encodes the representations of agents' historical motions and candidate route waypoints separately; 2) cross-modality level that models the cross-modal interactions among agents and their candidate route waypoints; 3) aggregating level that composites the second level latent of neighbor agents to the ego; 4) output level that queries the candidate route waypoints for the ego agent from the third stage, concatenating with selected information flow. Each level of encoding is detailed below. 

\emph{1) Dynamic Level:} Given input state $s_t$, dynamic level separately encodes the historical motions $\mathcal{M}_t$ and corresponding local candidate waypoints $\mathcal{K}_t$ for each vehicle. Hereby we introduce a single layer Transformer along the temporal axis for dynamic level encoding. 
\begin{equation}
    \label{e3}
    \operatorname{Trans^M_T}(\mathbf{X},\mathbf{M}) = \operatorname{MLP}(\operatorname{MaxPool}(\operatorname{MHA}(\mathbf{X},\mathbf{M}))),
\end{equation}
where the input $\mathbf{X} = \mathbf{Q}, \mathbf{K}, \mathbf{V}$ denoting the self-attention mechanism for temporal features, and max pooling is adopted to prevent over-fitting. 

To distill the temporal feature for candidate route waypoints, the vehicle embedding $\mathbf{Emb}$ is concatenated to differentiate the categorical feature between ego vehicle and neighbors, since their representations are then queried under different heuristics in \emph{Level} 2) and 4). The temporal Transformer for candidate route waypoints is represented as:
\begin{equation}
    \label{e3.2}
    \operatorname{Trans^P_T}(\mathbf{X},\mathbf{M}) = \operatorname{MLP}(\text{concat}(\operatorname{MHA}(\mathbf{X},\mathbf{M}),\mathbf{Emb})).
\end{equation}

We omitted the position embedding input since both $\mathcal{M}_t$ and $\mathcal{K}_t$ carry dense positional relations themselves. The output dynamics embedding can be represented as:
\begin{equation}
    \label{e4}
    D^i_{Mt} = \operatorname{Trans^{M}_T}(M^i_t, \mathbf{M}^i_M), \ d^i_{\mathbf{k}t} = \operatorname{Trans^{P}_T}(\mathbf{k}^{i,j}_t, \mathbf{M}^i_{kj}),
\end{equation}
where $i \in ego, 1, 2,\cdots, n$ stands for all vehicles (ego and neighbors from 1 to $n$), and $j \in 1, 2, \cdots$ is the number of candidate waypoints for each vehicle. $\operatorname{Trans^{M}_T}$ and $\operatorname{Trans^{L}_T}$ denote separated temporal Transformers, encoding historical motions and all candidate route waypoints of each vehicle $M^i_t$, $K^i_t = \{\mathbf{k}^1_t, \mathbf{k}^2_t, \cdots\}$ with temporal masks $\mathbf{M}^i_M, \mathbf{M}^i_K = \{\mathbf{M}^i_{\mathbf{k}1}, \mathbf{M}^i_{\mathbf{k}2}, \cdots\}$. The first stage outputs latent representations for vehicle motion dynamics $D^i_{Mt}$ and matching latent sets for candidate waypoints: $ D^i_{Kt} = \{d^{i,1}_{\mathbf{k}t}, d^{i,2}_{\mathbf{k}t}, \cdots\}$. 

\emph{2) Cross Modality Level:} The second stage seeks to abstract the interaction between neighboring vehicles' motion state $D^i_{Mt}$ and their local waypoint $D^i_{Kt}$. This level is modeled under the real-world heuristic that there is no cooperation among neighboring vehicles, hence the future waypoints are solely queried by their own historical motion dynamics. The cross-modality Transformer is designed along the axis of prediction numbers as follows:
\begin{equation}
\label{e5}
\operatorname{Trans_C}(\mathbf{X}, \mathbf{Y}, \mathbf{M}) = \operatorname{MLP}(\operatorname{MHA}(\mathbf{X}, \mathbf{Y}, \mathbf{M})),
\end{equation}
where cross-modality inputs $\mathbf{X}=\mathbf{Q}, \mathbf{Y}=\mathbf{K}, \mathbf{V}$ and mask $\mathbf{M}$. The cross-modality dynamic is represented as:
\begin{equation}
\label{e6}
C^i_t = \operatorname{Trans^C_C}(D^i_{Mt}, \ \text{concat}(D^i_{Kt}), \mathbf{M}^i_\mathbf{k}) + D^i_{Mt},
\end{equation}
where $i = 1,2,\cdots,n$. The residual connection is added to emphasize the dynamic $D^i_{Mt}$ of neighbor vehicles, while preserving the information flow in cases unavailable for $K^n_t$.

\emph{3) Aggregating Level:} The third stage focuses on the agent-level interactions of ego vehicle for decision-making with motion and map information flow for other neighbor vehicles. It is still assumed that in the scene, there is no interaction among neighbor vehicles. Thus in this level, the scene representation is encoded by aggregation with the ego dynamic itself $D^{ego}_{Mt}$ and the cross-modality output $\{C^1_t, \cdots, C^n_t\}$ for neighbor vehicles, the structure also follows the cross attention scheme:
\begin{equation}
    \label{e6}
    Ag_t=\operatorname{Trans^A_C}(D^{ego}_{Mt}, \ \text{concat}(D^{ego}_{Mt}, C^1_t, \cdots, C^n_t), \mathbf{M}_n).
\end{equation}

\emph{4) Output Level:} The output stage also counts the future heuristic for the ego vehicle, which should be determined through interaction awareness of neighbor vehicles. The intuition is to query the candidate route waypoints given ego dynamics and its interactions with surrounding neighbors which also have knowledge of their future intentions. Another cross-attention Transformer is designed to model this intention:
\begin{equation}
    \label{e7}
    h_t=\operatorname{Trans^E_C}(Ag_t, \ \text{concat}(D^{ego}_{Kt}), \mathbf{M}^{ego}_\mathbf{k}) + Ag_t.
\end{equation}

The final representation $h_t$ is also passed through a residual connection in the output level. It is a well-structured representation given all the interactions from different levels of the input state. 

\subsection{Soft actor-critic}
We implement Soft Actor-Critic (SAC) for decision-making given latent representation $h_t\leftarrow \operatorname{\Phi}(s_t)$ through the MST structure. SAC is a model-free off-policy RL method with state-of-the-art performance in continuous decision-making tasks \cite{b33}. Its objective maximizes both cumulative rewards and entropy with a temperature parameter $\alpha$ to regularize the exploration-exploitation process:
\begin{equation}
\label{e8}
\underset{s_{t}, a_{t} \sim \pi}{\mathbb{E}}[\Sigma_{t} \mathcal{R}\left(s_{t}, a_{t}\right) + \alpha \mathcal{H}(\pi(\cdot|s_t))].
\end{equation}

SAC simultaneously learns a stochastic policy network $\pi_\phi$ (actor) and double Q-function networks $Q_{\theta_1}, Q_{\theta_2}$ for variance reduction of the Q values (critic). Given the latent representation $h_t$, the Q-function networks are updated according to the mean Bellman squared error (MBSE) over MDP tuples $\tau=(s_t,a_t,r_t,s_{t+1}, \gamma)$ sampled from replay buffer $\mathcal{D}$:
\begin{equation}
\label{e9}
\mathcal{L}_{\text {c}}(\theta_i)=\underset{\tau \sim \mathcal{D}}{\mathbb{E}}\left[\left(Q_{\theta_i}(h_{t}, a_{t})-(r_{t}+\gamma y_Q)\right)^{2}\right],
\end{equation} 
where $i=1,2$, latent scene representations $h_t\leftarrow \operatorname{\Phi}(s_t)$ is acquired through MST. The temporal difference (TD) target $y_Q$ for soft Q-function update is given by MDP tuples $\tau$ above and actions $a^{\prime}$ sampled by current policy $\pi$:
\begin{equation}
\label{e10}
y_Q = \underset{a^{\prime}\sim\pi}{\mathbb{E}} \left[\min_{i=1,2}Q_{\bar{\theta}_i}(\bar{h}_{t+1}, a^{\prime})-\alpha \log \pi_{\phi}(a^{\prime}|h_{t+1}) \right],
\end{equation}
where $h_{t+1}\leftarrow \operatorname{\Phi}(s_{t+1}), \bar{h}_{t+1}\leftarrow \operatorname{\bar{\Phi}}(s_{t+1})$. In the critic, target Q-values and their representations $\bar{h}_{t+1}$ are acquired through the target network $\bar{\Phi}$ which is dynamically updated by Polyak averaging at each gradient step. The minimum Q-value from double Q networks $\bar{\theta}_{1,2}$,  is taken to prevent overestimating issues for state action values.

The actor policy network outputs the parameters (means and covariance $\mu,\Sigma\in\mathcal{A}$) of a multivariate Gaussian distribution $\mathcal{N}(\mu,\Sigma)$, each with dimensions of action space. The actor is updated by soft-Q minimization:
\begin{equation}
\label{e11}
\mathcal{L}_{a}(\phi)=-\underset{\tau \sim \mathcal{D}}{\mathbb{E}}\left[\min_{i=1,2}Q_{\theta}(h_{t}, a_t)-\alpha \log \pi_{\phi}(a_t|h_{t})\right],
\end{equation}
where the policy is taken from the state representation from the MST but stopping gradient backpropagation: $h_t\leftarrow sg(\operatorname{\Phi}(s_t))$. 
The parameter $\alpha$ controls the trade-off of the entropy term. is automatically tuned over the course of training 
\begin{figure}[htp]
    \centering
    \vspace{-0.2cm}
    \includegraphics[width=\linewidth]{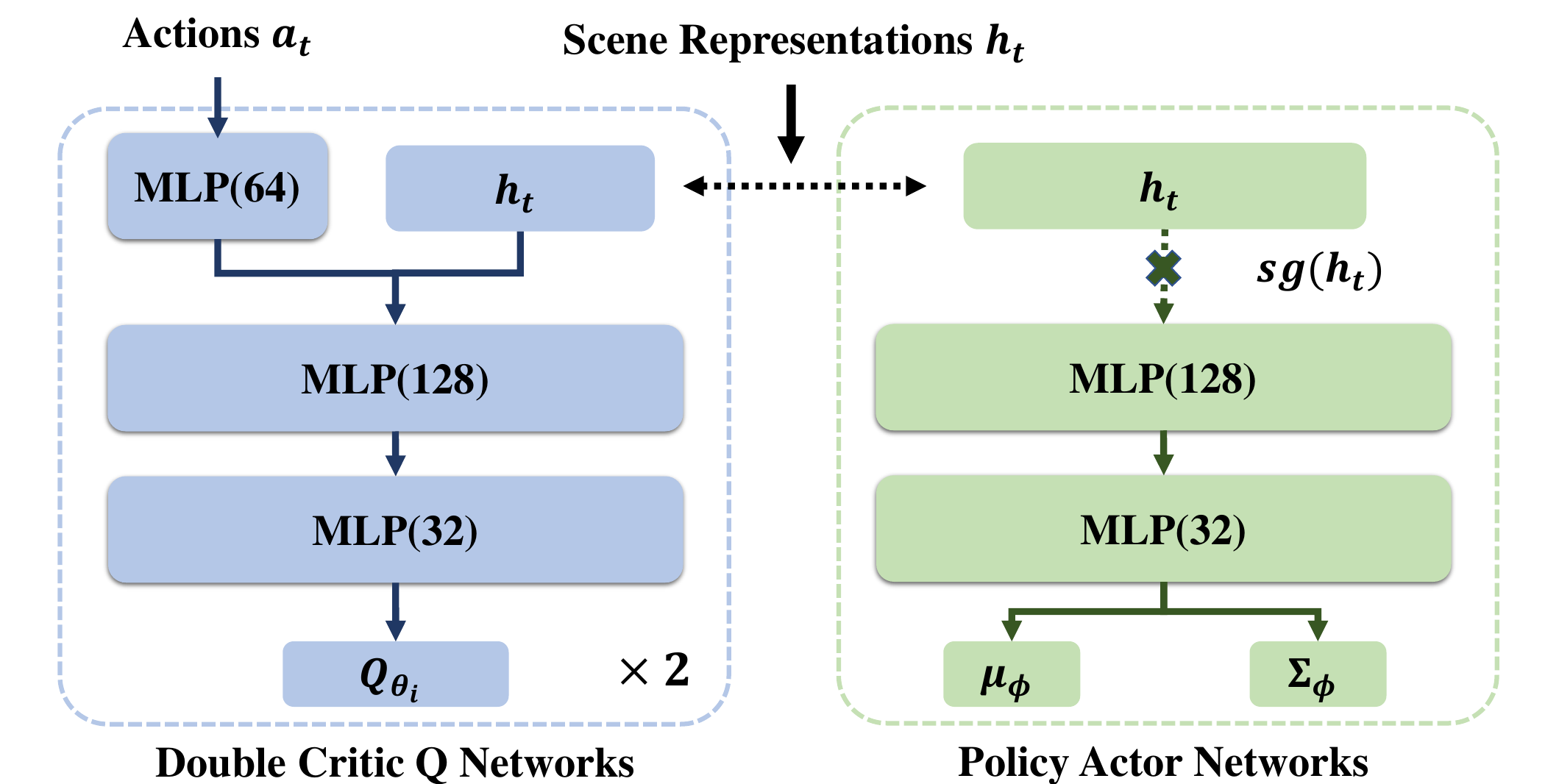}
    \caption{The structure of Actor-Critic networks. For continuous decision-making tasks, the policy is modeled by multivariate Gaussian with mean and covariance diagonal.}
    \label{fig:fig.4}
\end{figure}
\begin{figure*}[htp]
    \centering
    \includegraphics[width=0.8\linewidth]{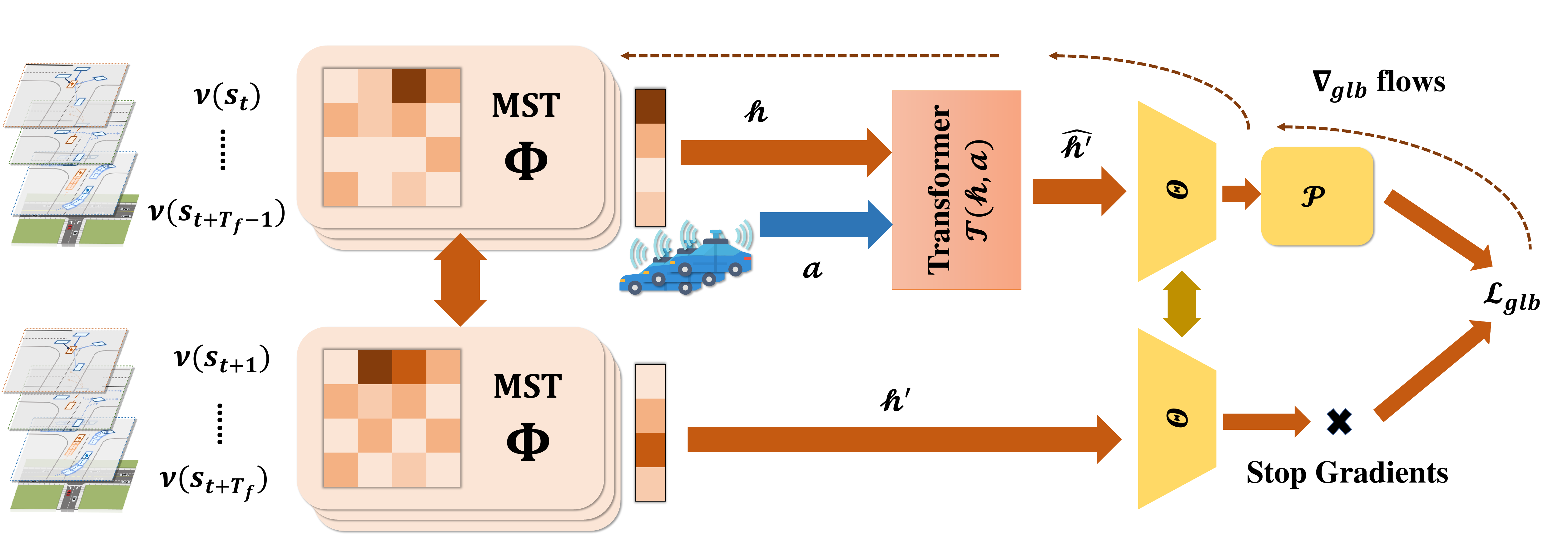}
    \caption{The computational flow of Sequential Latent Transformer. Sequential consistent features among future horizons $T_f$ are captured through this Siamese network structure. All augmented future states are encoded by MST $\Phi$. It then follows a sequential representation learning with representation transition model $\mathcal{T}$ and acquires projected similarity loss $\mathcal{L}_{glb}$ for this auxiliary task.  {Proof for SLT framework can be found at Theorem} \ref{theorem}}
    \vspace{-0.5cm}
    \label{fig:fig.5}
\end{figure*}
  according to \cite{b33}. Actions are sampled from the current policy $a_t\sim \pi_\phi(\cdot|\Phi(s_{t}))$ during training and take the mean of policy $a_t=\mu_\phi(\cdot|\Phi(s_{t}))$ during testing.

\subsection{Sequential Latent Transformer}
 {From a global perspective, the decision-making process can be improved by having prior knowledge of the decision consequences. This can be improved as: 1) global prediction distills the consecutive state-action pairs that reduce the exploration space; 2) the attention weights for route waypoints $\mathcal{K}_t$ can be implicitly guided by global predictions. 
Following the guideline, an auxiliary task should be designed to learn the latent representations $h_t$ predictable to the next-step state $h_{t+1}$ conditioned on current actions $a_t$ during training. It could be modeled as a forward environment transition:}
\begin{equation}
\label{e12}
{h^\prime}_{t+1} \leftarrow \mathcal{T}(h_{t+1}|h_{t}, a_{t}).
\end{equation}

 {Nonetheless, a single-step forward model confronts a series of problems: 1) representations are sequentially dependent so that the consequence may be led by a series of actions; 2) the highly dynamic nature of the driving scene requires distilling the sequential representations to combat the variances; 3) auto-encoder method accumulates great errors in sophisticated urban state reconstructions. Moreover, current modeling for auxiliary tasks lacks analytical originality. Therefore, a sequential latent transformer (SLT) is proposed as an auxiliary representation learning task during training. We also derive the theoretical proof for proposed pipelines.}

To deal with the first problem, the transition model in Eq. \ref{e12} should be modified to an auto-regressive structure to model the sequential relationships among scene representations and the corresponding actions under a future horizon $T_f$:
\begin{equation}
\label{e13}
{h^\prime}_{t+k} \leftarrow \mathcal{T}_{k\leq T_f}(h_{t+k} | h_{<t+k}, a_{<t+k}).
\end{equation}

For the second problem, a representation learning method is introduced. By computing the projected latent similarity among different aspects of the same state input through a Siamese network, the framework captures the invariant features from different aspects of inputs. The third problem is also solved since the method is reconstruction-free. To combat the variance issue, we follow the representation learning method \cite{b34}, which augments the input $\upsilon(s_t)$ to reduce the variance by averaging the same representation with different aspects of augmentations. It is also beneficial to decision-making since the Q-value over-estimation issues would be regularized with the same Q-value over different aspects of state inputs \cite{b29}.

\begin{figure}[tp]
    \vspace{-0.2cm}
    \centering
    \includegraphics[width=\linewidth]{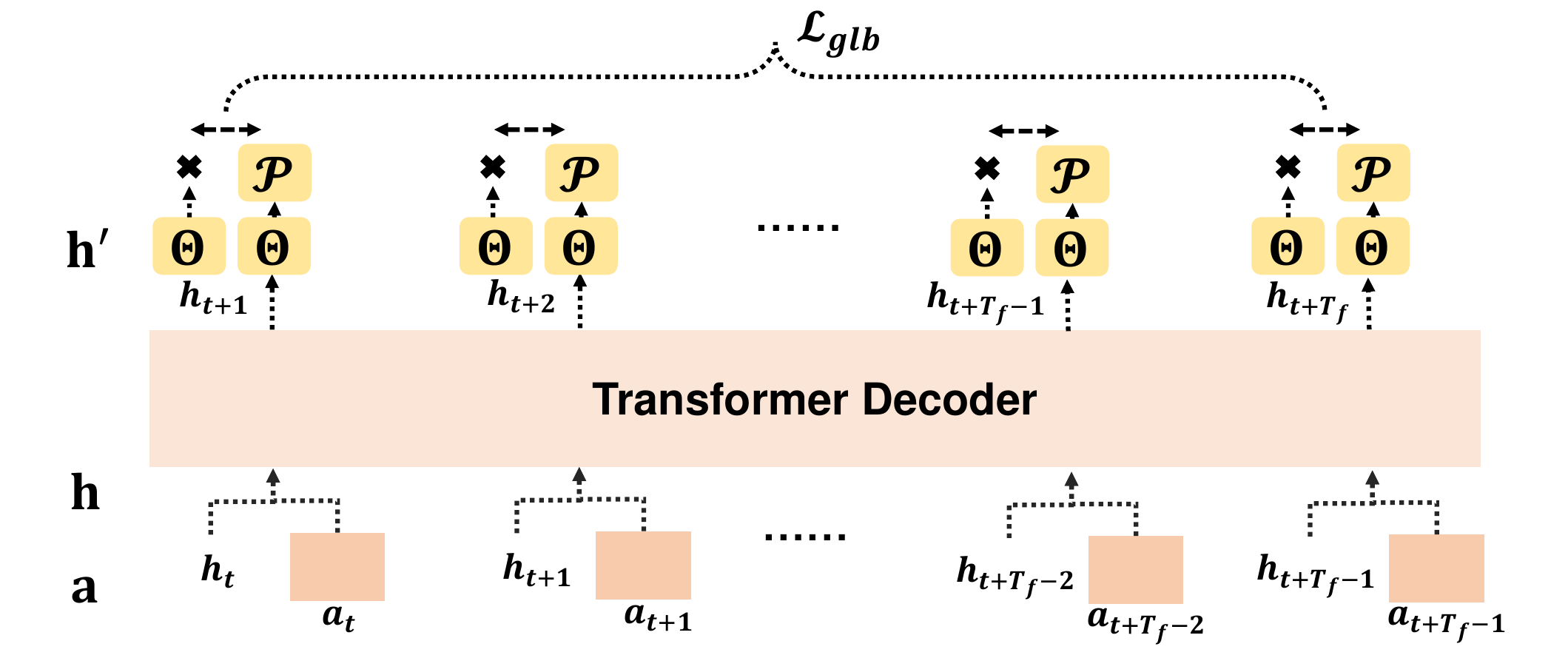}
    \vspace{-0.5cm}
    \caption{ {The structure of the Transformer-based transition model $\mathcal{T}$. It is a temporal Transformer decoder given concatenated input of $h_t, a_t$ for each timestep, and the output is representation predictions for the next step as an auto-regressive way for projections and predictions.}}    
    \label{fig:fig.6}
\end{figure}

\textbf{SLT Computational Framework:} Fig. \ref{fig:fig.5} shows the structure of SLT, which follows a SimSiam-style representation learning method \cite{b34}. Given the continuous state-action pairs over future horizon $T_f$: 
$$(\upsilon(s_t),a_t), (\upsilon(s_{t+{1}}),a_{t+{1}}), \cdots, (\upsilon(s_{t+{T_f}}),a_{t+{T_f}}),$$
then all augmented states would be encoded through MST: $h_t \leftarrow \Phi(\upsilon(s_t))$. Let $\mathbf{h}, \mathbf{a}$ denote the representations and actions sequences $h_{<{T_k}}, a_{<T_k}$, the future step representations can be gathered into $\mathbf{h^\prime} = h_{t:t+T_k}$. As shown in Fig. \ref{fig:fig.6}, the predicted future latent representations are acquired through an auto-regressive Transformer-based decoder: $\hat{\mathbf{h^\prime}}=\mathcal{T}(\mathbf{h}, \mathbf{a})$. Both $\mathbf{h^\prime}, \hat{\mathbf{h^\prime}}$ are projected with the predictor for similarity computation:
$z = sg(\Theta(\mathbf{h^\prime})), \hat{z}=\mathcal{P}(\Theta(\hat{\mathbf{h^\prime}}))$, where $\Theta$ and $\mathcal{P}$ are MLP projector and a linear predictor, respectively. Gradient flow $z$ is stopped for the target. The similarity loss for global predictions $\mathcal{L}_{glb}$ is updated simultaneously with RL decision-making. During inference, we strictly follow the general representation learning paradigms \cite{b28} which only reserve augmented $\Phi$ without updating the loss term. The implementation of SAC with the proposed Scene-Rep Transformer is given in Algorithm \ref{Ag}.

\textbf{Similarity Loss:} To prevent the representation collapse, cosine similarity is proposed as the similarity loss function \cite{b35}:
\begin{equation}
\label{e14}
\mathcal{L}_{glb}(z,\hat{z})= -\frac{z\cdot\hat{z}}{\|z\|\cdot\|\hat{z}\|}. 
\end{equation}
The loss is updated separately since all objectives do not share the same magnitude and may conflict with each other for optimization. Networks involving similarity loss computations are $\Phi$, $\mathcal{T}$, $\Theta$, and $\mathcal{P}$.

\textbf{State Augmentation:} The state augmentation $\upsilon$ given input state $s_t$ comprises two steps for driving scenarios. 1) The absolute coordinate is transformed to be Cartesian coordinate centered on the ego vehicle along its longitudinal axis to focus more on the local variations. 2) All points are processed with random rotations
 between $[-\frac{\pi}{2}, \frac{\pi}{2}]$ by ego center to build different views for the states.

\textbf{Data Collections:} During training, we collect the transition tuples $\tau$ into replay buffer, as well as maintaining a queue $\mathcal{D}^f$ to collect the corresponding future state-actor pairs $(\mathbf{s}, \mathbf{a}) = (s_{t:t+T_f}, a_{t:t+T_f})$ over the time horizon $T_f$.

 {\textbf{SLT Analytical Verification:} SLT aims to distill the core consistent future feature for predictable representations. Theoretically, it is objectified as maximizing the future information bottleneck for latent dynamic models $\mathcal{T}$: $\max \left[ I(\mathbf{h},\mathbf{s}|\mathbf{a})-\beta I(\mathbf{h},\mathbf{s}^i|\mathbf{a}) \right]$ 
, where $i$ is the sampled index }\cite{alemi2016deep}.  {Theoretical proof for SLT can be found at Theorem} \ref{theorem}

\begin{algorithm}[htb]
	\vspace{-\topsep}
	\caption{Soft Actor-Critic with Scene-Rep Transformer}
	\begin{algorithmic}[1] 
	\Require Initial entropy parameter $\alpha$, Polyak weight $\lambda$
	\State Initialize MST $\Phi$ and SLT $\mathcal{T},\Theta,\mathcal{P}$
	\State Initialize policy network $\phi$, Q value networks $\theta_{1,2}$
	\State Initialize target networks $\bar{\Phi}$, $\bar{\theta}_{1,2}\leftarrow \Phi,\theta_{1,2}$
	\State Initialize empty replay buffer $\mathcal{D}$, future queue $\mathcal{D}^f$
	\Repeat
        \State Sample ego vehicle action ${a}_{t} \sim \pi_\phi(\cdot|\Phi(s_{t}))$ 
		\State Interact with environment $\tau=(s_t,a_t,r_t,s_{t+1})$
		\State Store $\tau$ in deque $\mathcal{D}^f$
		\If {$T_{epi}>T_f$}
		    \State Store in buffer $\mathcal{D}$ with $(\tau,\mathbf{s},\mathbf{a})\sim\mathcal{D}^f$
		\EndIf
		\If {time to update}
		    \State Sample batch from replay buffer: $\tau,\mathbf{s},\mathbf{a}\sim\mathcal{D}$
		    \State State augmentation: $s_{t:t+T_f}=\upsilon(s_{t:t+T_f})$
		    \State \textbf{RL decision making:}
    		\State Critic loss $\mathcal{L}_{\text {c}}=\mathcal{L}_{\text {c}}(\theta_1)+\mathcal{L}_{\text {c}}(\theta_2)$ (Eq. (\ref{e9}))
    		\State Polyak weight $\lambda$ averaging for $\bar{\Phi},\bar{\theta}_{1,2}$
    		\State Actor loss $\mathcal{L}_{a}(\phi)$ (Eq. (\ref{e11}))
    		\State Update entropy parameter $\alpha$ 
    		\State \textbf{Predictive representation learning:}
    		\State Scene encoding by MST: $\mathbf{h}, \mathbf{h^\prime} \leftarrow \Phi(\mathbf{s})$
    		\State Predicted transitions $\hat{\mathbf{h^\prime}}=\mathcal{T}(\mathbf{h},\mathbf{a})$
    		\State Projections $z=sg(\Theta(\mathbf{h^\prime})),\hat{z}=\mathcal{P}(\Theta(\hat{\mathbf{h^\prime}}))$
    		\State Similarity loss $\mathcal{L}_{glb}(z,\hat{z})$ (Eq. \ref{e14})
    		\State Update $\mathcal{L}_{\text {c}},\mathcal{L}_{a}(\phi),\mathcal{L}_{glb}$ with separate optimizer
    	\EndIf
    	\If {$s_{t+1}$ is terminal}
    	\State Reset environment, $T_{epi}=0$
    	\State Update rest transitions in $\mathcal{D}^f$ to $\mathcal{D}$
    	\EndIf
    \Until{Convergence}
	\end{algorithmic}
	\vspace{-\topsep}
	\label{Ag}
\end{algorithm}
\vspace{-0.2cm}
\section{Experiments}
 {To validate the decision-making performance of Scene-Rep Transformer under realistic scenario characteristics, we introduce two realistic simulation platforms, namely CARLA }\cite{dosovitskiy2017carla}  {and SMARTS} \cite{b32}  {for comprehensive verification.  In CARLA we focus on verification in hyper-realistic dynamics and urban elements. In SMARTS the simulation features hyper-realistic interactions and stochastic traffic flows.}
\subsection{Driving Scenarios}
\begin{figure}[htp]
    \centering
    \includegraphics[width=\linewidth]{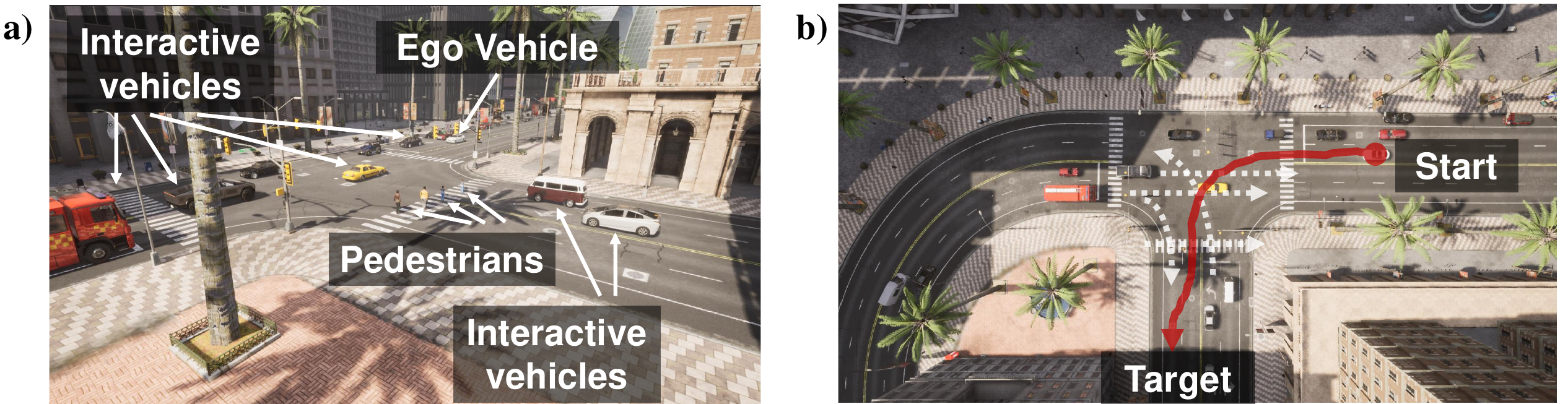}
    \caption{ {The interactive scenario in CARLA \texttt{Town-10}. a) The devised left-turn scenario with both vehicles and pedestrians; b) Bird's eye view of the task settings. The red line indicates an example ego trajectory. The white lines denote the possible interactive trajectories for social actors (vehicles and pedestrians).}  }  
    \label{fig:carla}
\end{figure}

 {In order to thoroughly assess the decision-making efficacy within real-world dynamics and urban-authentic scenarios, we have incorporated the CARLA simulator for validation purposes. CARLA offers hyper-realistic physics and dynamics, providing built-in AI Agents for traffic behaviors to each participant. In our experiment, we select a representative urban scenario in CARLA \texttt{Town-10}. As shown in Fig.}\ref{fig:carla},  {the ego autonomous vehicle is assigned with a demanding unsignalized urban intersection for decision-making validations. 

In this task, the ego vehicle is mandated to contend with different types of participants shown in Fig.}\ref{fig:carla}  {a). Each participant is controlled by AI Agents with randomly assigned behaviors (patient, aggressive, moderate) during interactions. For task completions shown in Fig.}\ref{fig:carla}  {b), the ego vehicle on the side road should first manage to make a left turn. It will confront interactive traffic involving: heading traffic from two opposite lanes, left-turn traffic from the left side lane, and pedestrians traversing the crosswalks. Moreover, the ego vehicle is tasked to perform at least one lane change, avoiding the right-turn traffic to reach the destination area. This selected scenario examines interactive driving decisions for avoiding, turning, and lane-changing under realistic dynamics and diversified urban traffic. }

 {Interaction awareness stands as a core contribution of the Scene-Rep Transformer. Consequently, it becomes imperative to confirm its decision-making ability within the realm of practical traffic interactions, encompassing a wider array of stochastic behaviors and traffic dynamics. To this end, we are committed to advancing our verification through the simulation platform within the SMARTS simulator for real-world traffic flow and interaction simulations.}

 { To manifest the decision performance under realistic interactions and actor behaviors, we 
  choose three challenging scenarios in official SMARTS platforms. SMARTS provides hyper-realistic interactions and traffic }
\begin{figure}[htp]
    \centering
    \includegraphics[width=\linewidth]{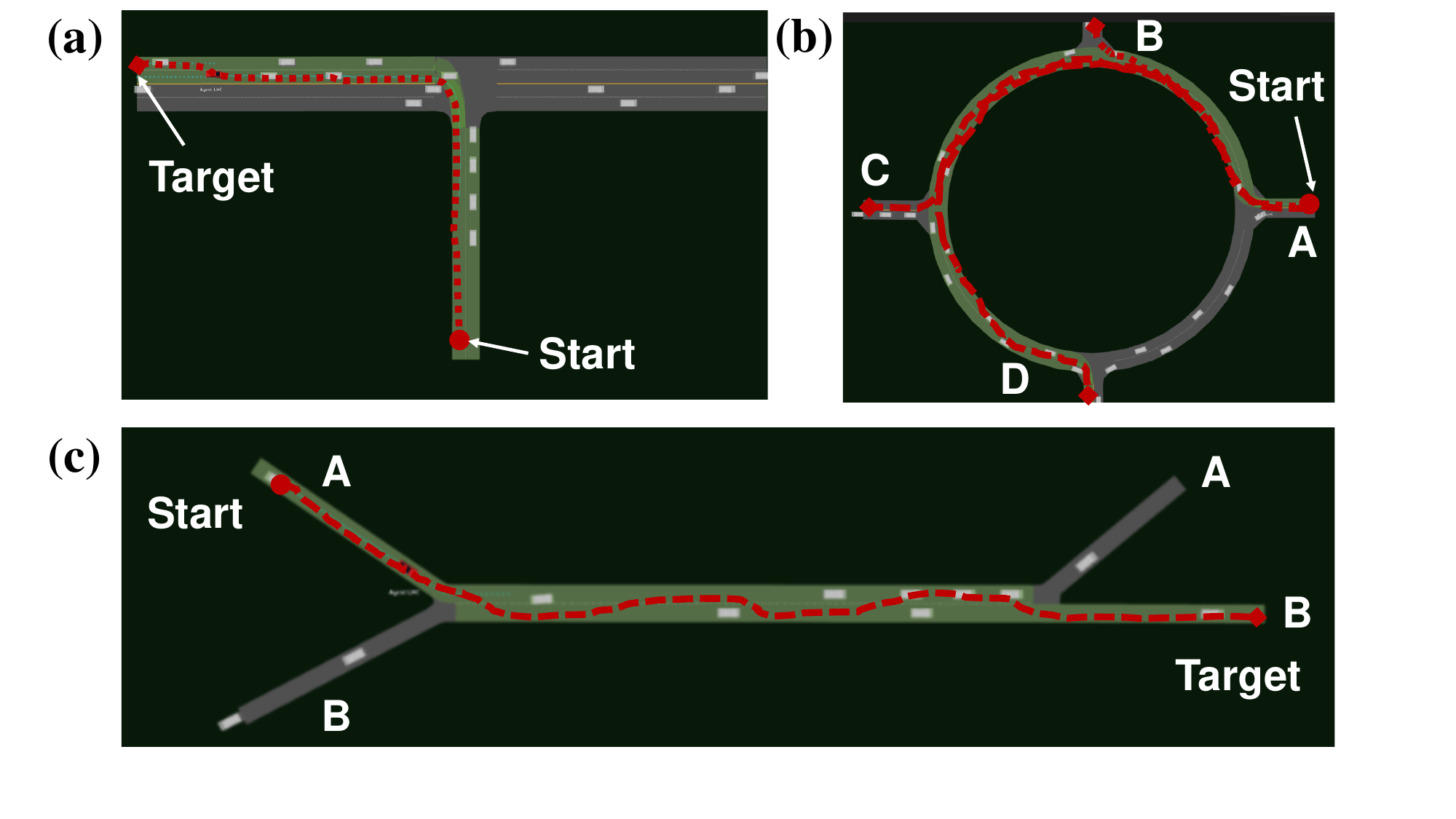}
    \caption{The designed urban experimental scenarios in SMARTS simulator platform. a) Unprotected left turn;  b) Roundabout; c) Double merging. The green color renders the viable area for the ego vehicle to reach the target. The red dots offer an example from the starting route to the task target}    
    \label{fig:fig.7}
\end{figure}
 {flows under different granularity. It further offers stochastic actor behaviors and imperfections based on a physics engine that is closer to reality. As shown in Fig. }\ref{fig:fig.7}. 
 {The scenarios are chosen for validating our framework from different aspects of realistic interactions and driving behaviors.}

\textbf{a) Unprotected left turn:} The scenario is built upon the urban unsignalized T-intersection with heavy traffic. The autonomous driving vehicle is required to complete a left turn task without traffic signal protection, and thus the traffic flow would not stop. Starting from the 2-lane single-way minor road, the ego vehicle is required to make smart decisions across the ongoing traffic flows onto the rightmost lane on the 4-lane two-way major road. This scenario focuses on maneuvers for smart traffic flow crossing.

\textbf{b) Double Merging:} The scenario faced is a 2-lane merging road segment with double entrances and exits. Note that the designed traffic flow is heavier than scenario a) since for each entrance, the traffic flow is assigned for all exits. The autonomous vehicle is required to merge from entrance A to exit B. It leaves a higher requirement for the ego vehicle to learn lane-changing focused maneuvers to finish the task.

\textbf{c) Roundabout:} A general city roundabout with 4 exits is designed with dense traffic flows. Entranced in crossing A, the autonomous vehicle is devised with 3 tasks with increasing difficulties to exits B, C, and D. It involves more complex decisions and longer ranges to complete the tasks.

 {
 In order to ensure objective and practical validations, we align the episodic settings for CARLA and SMARTS. The starting point is assigned randomly along the starting route, while the target is the endpoint for a certain route. Resetting of simulations occurs under the following circumstances: 1) the ego vehicle reaches the target; 2) the episodic step exceeds the maximum timesteps limit; 3) the ego vehicle collides with other objects or drives off the routes. To establish practical and realistic simulations, both scenarios in CARLA and SMARTS are devised from official configurations. In the CARLA scenario, participants are spawned randomly in time and installed with built-in AI Agents to exhibit assorted behaviors, chosen arbitrarily. For SMARTS simulations, each scenario encompasses an official default setting shown in Table} \ref{table_1}.  {In macroscopic granularity, the entire scenario is configured by dense traffic level. Based on the default value for realistic behaviors, actors are sampled for more stochastic and realistic behaviors over CARLA. The simulation interval is set to 0.1s. }

\subsection{Decision-making process}
The designed state space, action space, and reward function are listed below.

\textbf{1) State $s$:} State input $s_t = [\mathcal{M}_t; \mathcal{K}_t]$ has two different modalities. For historical motion, its state space $\mathcal{S}_\mathcal{M} \in [n+1, T_h, 5]$ represents the number of vehicles (ego and neighbors), historical steps, and feature dimensions. For local candidate route waypoints, its state space is $\mathcal{S}_\mathcal{K} = [n+1, N_\mathbf{k}, T_\mathcal{K}, 3]$, where the second dimension indicates the maximum number of local candidate routes $\mathbf{k}$. It can be adjusted according to the maximum lanes for different road structures.

\textbf{2) Action $a$:} Rather than end-to-end control for acceleration and steering, herein we combine the strategic-level decisions to be executed by the downstream controller. The decision-making actions $a_t=[V_t,L_t]$ comprise two-dimensional vectors for target speed $V_t$ and lane changing $L_t$. The former is a continuous output $V_t\in[0,V_{max}]$ range from 0 to a maximum speed limit $V_{max}$. The latter is a discrete action $L_t\in\{-1,0,1\}$, where $L_t=\pm1$ indicates a left/right lane changing, while $L_t=0$ means to keep the lane. Following the decision commands, a built-in motion planner from the SMARTS simulator \cite{b32} would provide corresponding routes, then the controller is employed to execute lane-follow controls (both lateral and longitudinal) on the ego vehicle. Note that a complete lane-changing process will take multiple steps for $L_t=\pm1$, and otherwise will be corrected back to the current lane by the controller. Similar processes are also followed in the CARLA simulator. Following the action commands, searched and processed motion-feasible planning routes are employed for subsequent control instead of raw points provided in CARLA.

To model the hybrid actions with continuous and discrete spaces, we introduce a normalized multivariate Gaussian distribution with 2 dimensions. For $V_t\in[0, V_{max}]$, it would be normalized to $[-1, 1]$. For $L_t$, we discretize the second dimension of the Gaussian to 3 equal-sized bins ($[-1, -\frac{1}{3}]$ for $L_t=-1$, $[-\frac{1}{3}, \frac{1}{3}]$ for $L_t=0$, and $[\frac{1}{3},1]$ for $L_t=1$). We choose this setting because 1) the target speed should be precise as it is critical for safety; 2) speed and lane-changing decision actions are somewhat correlated.

\textbf{3) Reward function $r(s,a)$:} To emphasize the framework contributions for the proposed Scene-Rep Transformer, we choose a simple and sparse goal-based reward as the reward function:
\begin{equation}
\label{e15}
r_{t} = r_{target} + r_{penalty},
\end{equation}
where $r_{target}=1$ is an indicator if the ego vehicle reaches the target, and $r_{penalty}=-1$ if the ego vehicle collides with others or drives out of the viable routes. The simple reward remains zero in other cases. 
 {For a thorough validation, our experiment also incorporates two denoted reward functions} \cite{b6,wang2020learning}  {for autonomous driving decision-making. Results can be found at\footnote{\change{{\href{https://github.com/georgeliu233/Scene-Rep-Transformer}{https://github.com/georgeliu233/Scene-Rep-Transformer}}}}. We also expect enhanced results by reward shaping.}

\subsection{Evaluation Metrics}
To make a fair evaluation of the performance of the proposed framework under different scenarios, we use the average success rate during training as the primary evaluation metric. During the test stage, we use additional metrics, i.e., the collision rate and stagnation rate.

\begin{itemize}
    \item \textbf{Success rate (Succ.\%):} It quantifies the percentage of episodes that the ego vehicle successfully reaches the target. During testing, we measure the success rate over the total testing episodes. For training cases, the success rate is measured over the last 20 episodes.
    \item \textbf{Collision rate (Coll.\%):} It counts the percentage of episodes in which the ego vehicle collides with others, which is a critical indicator for safety performance.
    \item \textbf{Stagnation (Stag.\%):} It counts the proportion of episodes that the ego vehicle stays still and exceeds the maximum time limit, which indicates the degree of conservativeness. 
\end{itemize}

\subsection{Comparison baselines}
To make a comprehensive evaluation of the proposed approach, we compare our SAC with the Scene-Rep Transformer with other existing methods for scene representations and decision-making. The baseline methods are:

1) \textbf{Data-regularized Q-learning (DrQ)} \cite{b29}: A state-of-the-art method for image-based RL (Dr-SAC for continuous action settings). Stacked rasterized images are used for representing the scene with roads and vehicles over consecutive time steps. The images would also go through random augmentations for regularization. It employs a convolutional neural network (CNN) for encoding the state and soft actor-critic (SAC) algorithm for decision-making.

2) \textbf{Soft Actor Critic (SAC)} \cite{b33}: An RL baseline method for decision-making without the proposed scene understanding framework. It takes the same representations for the state input but uses LSTM for encoding and aggregates the information from each step.

3) \textbf{Proximal Policy Optimization (PPO)} \cite{b36}: A state-of-the-art on-policy approach for decision-making. It improves the RL policy considering the constraints to stay close to the last updated policy using surrogate actor-critic objectives within an episodic horizon. We take the most common image-based scheme as input (the same as baseline 1)).

4) \textbf{Rule-based Driver Model (RDM)} \cite{b32}: Implementation with the same behavior model as neighbor vehicles, but with deterministic settings and behaviors without imperfections. 

\textbf{5) Decision Transformer (DT) \cite{chen2021decision}}: A strong baseline of Transformer's implementation for RL decision making. It models the RL to a sequential learning of MDP sequences via a Transformer decoder.

\subsection{Implementation details}
Our proposed approach and other baseline methods (except for RDM) are all trained for 100,000 steps in each urban scenario, and each experiment is conducted 5 times with different random seeds. For DT we log-replayed 500 successful MDP trajectories for each scenario and trained following its original implementation. The neural networks for all methods are trained on a single NVIDIA RTX 2080 Ti GPU using Tensorflow and Adam optimizer with a learning rate of $1e^{-4}$, and the training process for a method in a single scenario takes roughly 2 hours. The parameters related to the experiment are listed in Table \ref{table_2}. For each method, we take the policy network with the highest success rate during training for the subsequent testing phase.

\section{Results and Discussions}
\subsection{Training results}

\begin{figure*}[ht]
    \centering
    \includegraphics[width=\linewidth]{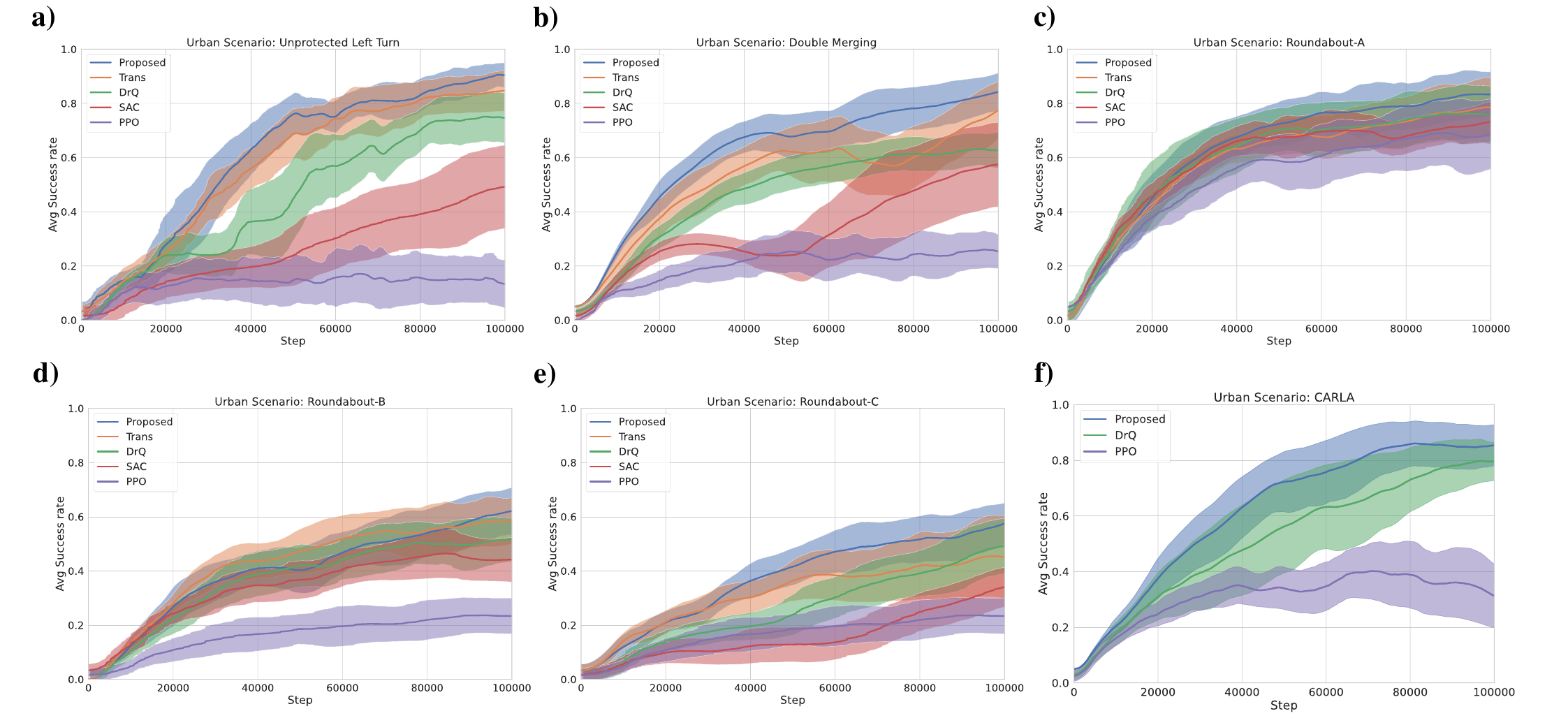}
    \vspace{-0.6cm}
    \caption{The training process of our framework with baseline methods quantified by average training success rate in different urban scenarios: a) Unprotected left turn; b) Double merge; c) Roundabout-A: ego vehicle to reach Exit B; d) Roundabout-B: ego vehicle to reach Exit C; e) Roundabout-C: ego vehicle to reach Exit D; f) Interactive urban scenario in CARLA. The blue curve (\textbf{Proposed}) indicates the Scene-Rep Transformer, and the orange curve (\textbf{Trans}) shows the ablative training performance with only MST. The other three training baselines are displayed in green (\textbf{DrQ}), red (\textbf{SAC}), and violet (\textbf{PPO}).}
    \vspace{-0.5cm}
    \label{fig:fig.8}
\end{figure*}

We first evaluate the training performance of the Scene-Rep Transformer together with the comparison baselines. The training results are displayed by urban scenarios, where each training curve represents the mean (solid line) and standard deviation error band of certain baselines for the average training success rate. The results are recorded every 200 steps and smoothed by exponentially moving average (EMA=0.99). Fig. \ref{fig:fig.8} shows the training results of different methods in all scenarios, as well as the ablated method (SAC with MST). 

The results show that SAC with the proposed Scene-Rep Transformer achieves the best training performance in all five urban scenarios with diverse tasks. The training success rate is boosted in all scenarios, especially in the more demanding urban scenes such as roundabout-C, unprotected left turn, and double merging. The sample efficiency of the proposed method is also enhanced, displaying faster convergence in fewer training steps. The results suggest that our proposed Scene-Rep Transformer can quickly adapt to different scenarios and is capable of reaching the $70\%-80\%$ of the final success rate in merely $50\%$ of the overall training steps. Here, we give some detailed explanations about the superior performance of our proposed method compared to the baseline methods.

\textbf{Compared with on-policy RL:} PPO can adapt well with fast convergence in simpler tasks such as roundabout-A (Fig. \ref{fig:fig.8}c), resulting in 0.7 of the average success rate and convergence only in 40k training steps. But it performs worse when encountering more complex traffic and longer mileage. The training curve remains flat after the initial policy improvement (20k-30k).  {It is partly due to the sensitivity of on-policy learning for reward and parameters. This can lead to data insufficiency and the need for more intensive fine-tuning when dealing with stochastic traffic flow and long-range maneuvers.}

\textbf{Compared with off-policy RL:} SAC driving policy can show satisfying stochasticity in the driving scenarios. SAC with the same state inputs as our proposed method shows a significant boost for the final success rate (0.5-0.73) compared with PPO. However, SAC leaves a long time for convergence and instability, reflected in larger variance across different trials ($\sigma>0.1$) (see Fig. \ref{fig:fig.8}a, b). This is mainly caused by a deficiency in state action pairs sampling for Q-value estimation or state representations encoding.

\textbf{Compared with efficient RL:} DrQ improves the success rate from $10.5\%$ to $49\%$ compared with the SAC baseline. It utilizes the data augmentation technique so that variance for Q-value estimation can be quickly converged as it provides more aspects of states for the same Q-value. It shows a faster convergence for 40k-50k training steps to reach the same performance level by the SAC baseline. However, the performance is still limited by the rasterized image inputs and CNN structure, because the image cannot explicitly model the interactions among vehicles. Thus, its final performance is close to the upper error band of the SAC baseline in highly interactive scenarios such as the merging cases (Fig. \ref{fig:fig.8}b).

\textbf{Compared with only MST:} SAC with MST improves the success rate in all scenarios on the basis of the SAC baseline, as well as the efficiency of convergence to $50\%-66\%$ of the training steps to reach the same performance of DrQ baseline. This also suggests that the SLT in the framework can keep the consistency among sequential driving scene and their decisions. Thus the proposed method results in better performance than only MST, from $84\%$ (Fig. \ref{fig:fig.8}a) in left turn scenario to $23\%$ (Fig. \ref{fig:fig.8}c) in the simplest roundabout-A case. A greatly improved convergence is also an advantage of the framework, taking $40\%-60\%$ of the training steps to reach the best performance of the DrQ method.

 {\textbf{Performances in CARLA:} Similar training performances are also reported in the CARLA environment. It further validates the learning ability under realistic urban scenarios with different traffic participants and real dynamics. Due to different image settings, here we directly employ MST as the scene encoding for all baselines. As a result, performance is improved of DrQ and PPO, which also manifests the effectiveness of multi-stage encoding.}


In summary, the performance improvement of the Scene-Rep Transformer during training could originate from three aspects. First, we incorporate the multi-stage Transformer structure (MST) to represent the interactions among the ego vehicle and its neighbors, as well as the adaptive fusion of local candidate route waypoints to represent the local intentions of the ego and neighbors for decision-making. Second, the SAC algorithm is adopted for decision making which satisfies the scholastic nature of urban scenarios. Finally, the sample efficiency is further elevated through the sequential latent transformer (SLT). The state-action exploration space is reduced by the guidance of global predictive knowledge that only retains consistent information among sequential future state-action pairs in different urban scenarios, and thus convergences are boosted under limited training steps.

\subsection{Testing results}
\label{testing_sec}
The testing scenarios are constructed with the same road networks but different settings for parameters in Table \ref{table_1} in terms of randomness and random seeds. The rule-based method (RDM) is also introduced during testing in comparison with other learning-based decision-making methods. Each of the following baseline and the Scene-Rep Transformer is tested for 50 episodes under each urban scenario. Statistics for three evaluation metrics (success rate, collision rate, and stagnation) during testing are presented in Table \ref{tab:tab2}. Note that the three metrics do not necessarily add up to 100\%, as the ego vehicle agent may reach the wrong destination occasionally.  

\begin{table*}[htbp]
\caption{Summary of the test results of five testing urban scenarios with three evaluation metrics}
\centering
\resizebox{\textwidth}{18mm}{
\begin{tabular}{c|ccc|ccc|ccc|ccc|ccc}
\toprule
\multirow{2}{*}{} & \multicolumn{3}{c}{\textbf{Left Turn}}  & \multicolumn{3}{c}{\textbf{Double Merging}} & \multicolumn{3}{c}{\textbf{Roundabout-A}} & \multicolumn{3}{c}{\textbf{Roundabout-B}} & \multicolumn{3}{c}{\textbf{Roundabout-C}} \\\cmidrule(l){2-16}
         & Succ.\% & Coll.\% & Stag.\% & Succ.\%& Coll.\% & Stag.\%& Succ.\%& Coll.\%& Stag.\%& Succ.\%& Coll.\% & Stag.\%& Succ.\%& Coll.\%& Stag.\% \\\toprule
RDM      & 2        & 54       & 44       & 0          & 100         & 0         & 68         & 30       & 0        & 2         & 98       & 0        & 0         & 100        & 0        \\
PPO      & 36       & 50       & 10       & 36         & 64        & 0         & 66        & 34       & 0        & 42        & 58       & 0        & 38        & 50       & 12       \\
SAC      & 68       & 28       & 0        & 62         & 22        & 0         & 76        & 24       & 0        & 48        & 52       & 0        & 46        & 48       & 6        \\
DrQ      & 78       & 20       & 0        & 76         & 14        & 0         & 80        & 20       & 0        & 72        & 28       & 0        & 68        & 30       & 2        \\
DT     & 66       & 32       & 0        & 70         & 30        & 0         & 76        & 22       & 0        & 68        & 32       & 0        & 66        & 30       & 0        \\
MST     & 88       & 12       & 0        & 92         & 4         & 0         & 84        & 16       & 0        & 76        & 24       & 0        & 66        & 34       & 0        \\\midrule
\rowcolor{mgray}
\textbf{Proposed} & 94       & 4        & 0        & 96         & 2         & 0         & 88        & 12       & 0        & 82        & 18       & 0        & 76        & 24       & 0 \\\bottomrule
\end{tabular}}
\vspace{-0.5cm}
\label{tab:tab2}
\end{table*}

The testing results in Table \ref{tab:tab2} are in accordance with the training outcomes. Our proposed Scene-Rep Transformer RL framework achieves the best performance in all five urban scenarios. The proposed method brings about the highest test success rate with much fewer failure cases by collisions. Apart from the performance brought by global predictive knowledge, the capability of local predictions aided with Transformer is also verified in the MST baseline, resulting in an overall improvement compared with the DrQ baseline. However, the MST is still limited 
\begin{table}[tbp]
\caption{Test results of task completion time (s)}
\centering
\setlength{\tabcolsep}{1mm}{

\begin{tabular}{c|ccccc}
\toprule
\multirow{2}{*}{} & \multicolumn{1}{c}{\multirow{2}{*}{\textbf{Left Turn}}} & \multicolumn{1}{c}{\multirow{2}{*}{\textbf{Double Merging}}} & \multicolumn{3}{c}{\textbf{Roundabout}}                                        \\\cmidrule(l){4-6} 
                  & \multicolumn{1}{c}{}                           & \multicolumn{1}{c}{}                                & \multicolumn{1}{c}{\textbf{A}} & \multicolumn{1}{c}{\textbf{B}} & \multicolumn{1}{c}{\textbf{C}} \\\midrule
PPO               & 36.4$\pm$6.7                                          & 36.3$\pm$3.0                                                & 23.0$\pm$4.9                 & 43.5$\pm$1.8                 & 63.5$\pm$1.7                 \\
SAC               & 19.2$\pm$0.5                                         & 34.9$\pm$1.9                                           & 25.2$\pm$1.3                 & 44.0$\pm$1.1                & 60.7$\pm$1.4                 \\
DrQ               & 18.2$\pm$0.4                                          & 34.9$\pm$1.7                                               & \textbf{22.4$\pm$0.3}                 & 40.3$\pm$0.5                 & 57.4$\pm$1.8                 \\

DT            & 26.3$\pm$4.6                                          & 26.7$\pm$5.6                                               & 25.5$\pm$4.8                & 34.4$\pm$4.2                 & \textbf{50.1$\pm$6.7}                \\

MST               & 17.3$\pm$2.9                                          & 30.9$\pm$2.5                                              & 28.3$\pm$3.3                & 35.3$\pm$1.9                 & 58.5$\pm$2.2                 \\\midrule
\textbf{Proposed}          & \textbf{12.5$\pm$0.4}                                          & \textbf{28.6$\pm$0.6}                                               & 24.5$\pm$1.2                 & \textbf{33.7$\pm$1.2}                 & 56.6$\pm$2.2  \\\bottomrule               
\end{tabular}

}
\label{table_4}
\end{table}
by its sole guidance for state space compared with the guidance of both state action space sequentially through global predictions. Thus, for urban scenarios that require long-term driving maneuvers (Roundabout-B and C in Table \ref{tab:tab2}) that complicate the exploration space, the performance of MST is similar to DrQ as they both merely place emphasis on the state space. The performances of SAC and PPO are alike in their training results, mainly due to their limitations illustrated in the previous section. Coinciding with previous results, DT performs better in long-term tasks (R-C). Still, DT is compromised by interactive and short-term task (Left turn), due to the verbosity and lacks of interaction modeling. Notably, the rule-based driver model displays terrible performance in testing, due to its inability to cope with stochastic traffic flows and collides with aggressive vehicles to cut in the ego vehicle's driving routes.

For the metric regarding driving efficiency (Stag.\%), learning-based baselines show considerable performance with scarce cases of stagnation for the ego vehicle. Cases over time step limits are mainly concentrated in Roundabout-C scenarios for PPO and SAC, as it requires longer task completion time, and may stop at a traffic jam in such scenarios. The highest stagnation for inefficiency occurs from the rule-based model in the unprotected left turn. As a rule-based driving agent suffers from over-conservative behaviors when faced with dense traffic from the main road, the ego vehicle will remain to wait at the T-intersection until it reaches the time limit. The testing results clearly indicate that the rule-based driving model shows poor performance when encountering urban scenarios with more sophisticated and stochastic settings.   

To further investigate the driving efficiency of learning-based baselines with Scene-Rep Transformer, the mean and standard deviation of the completion time for successful trials during testing episodes are computed for each urban scenario (Table \ref{table_4}). The results reflect the improving traveling efficiency of the proposed method under the majority of urban scenarios. 
\begin{table}[tbp]
\caption{Test results in CARLA simulations}
\centering
\setlength{\tabcolsep}{1.2mm}{
\begin{tabular}{c|cccc}
\toprule
   & Succ.\% & Coll.\% & Stag.\% & Completion time (s) \\\midrule
   
PPO                                                       & 56                                                & 38                & 6                 & 22.6$\pm$1.7                 \\
DrQ                                                        & 84                                               & 10                 & 4                 & \textbf{21.3$\pm$1.4}                 \\

DT                                                    & 70                                              & 18                & 8                 & 22.6$\pm$1.5                \\

\midrule
\textbf{Proposed}                                    & \textbf{88}                                              & \textbf{6}                 & 2                & 22.9$\pm$1.6  \\\bottomrule               
\end{tabular}}
\label{table_carla}
\end{table}
It takes a longer time for PPO and SAC agents to finish the  
 task. They both show tentative behaviors to strictly follow the routine maneuvers for PPO or act with randomness due to the bias of local predictions for SAC. DrQ delivers better traveling efficiency and lower variance owing to the data regularization mechanism during training. This is further extended by the proposed Scene-Rep Transformer through the data augmentation for multi-step future pairs, leading to similar features. 

 {To validate the hyper-realistic performances for dynamics and actor diversity, we conducted testing in CARLA with the same settings and metrics as used in SMARTS. 
The results showed a similar trend for all baselines, with a slight drop in performance compared with the left-turn scenario in SMARTS. This decrease is partially attributed to the more complex dynamics and diverse range of traffic participants (i.e. pedestrians and social vehicles), which were not specifically modeled for their types in our framework, and directly serve in state inputs. It is important to note that the performance drop in CARLA was greater than in SMARTS. This is likely due to occasional dead-lock effects experienced by the interactions of other participants in CARLA.}

\subsection{Ablation study}
\begin{figure*}[htbp]
    \centering
    \vspace{-0.2cm}
    \includegraphics[width=\linewidth]{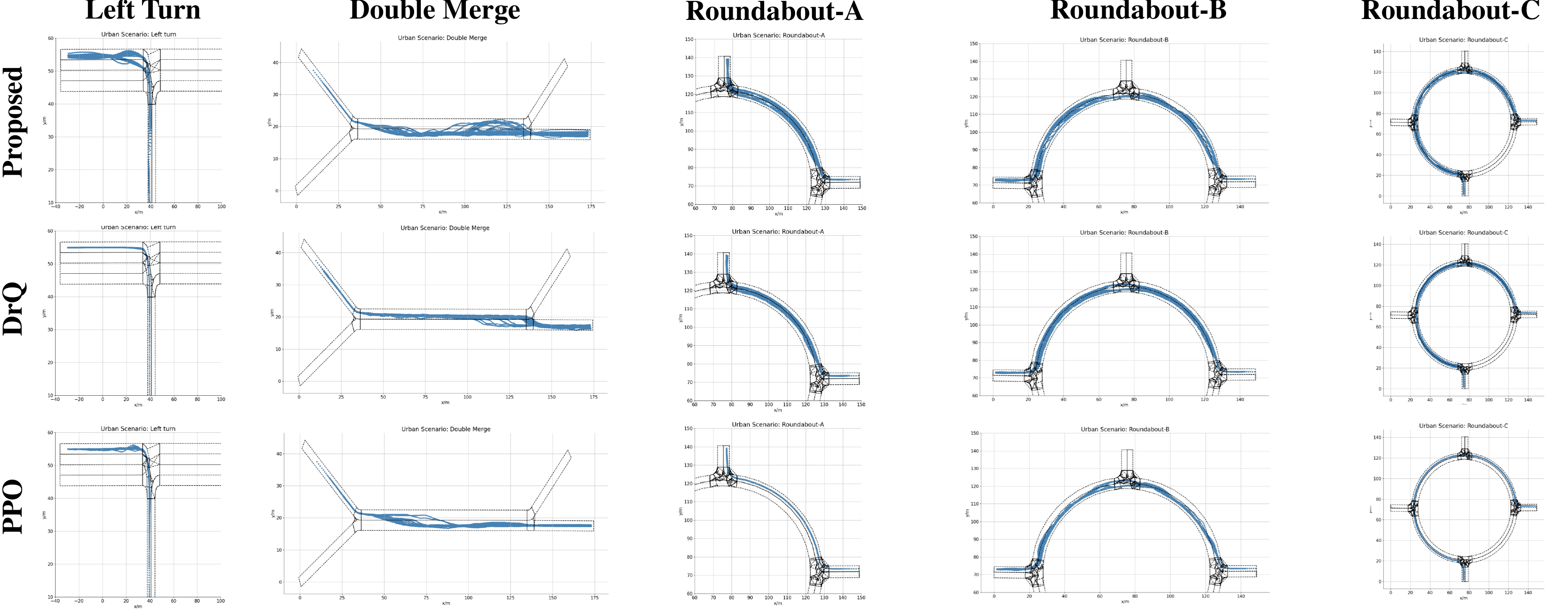}
    \vspace{-0.5cm}
    \caption{The trajectories of the autonomous vehicle that successfully completes the given tasks during the testing stage. The column indicates the five different urban scenarios while the row is for comparison of our method with DrQ and PPO. The trajectory distribution implies the attention predictive information in Scene-Rep Transformer endows more diversified long-term policies (trajectories) for task completion.}
    \label{fig:fig.9}
    \vspace{-0.2cm}
\end{figure*}
To explore the roles of predictive representation learning and candidate route waypoints in Scene-Rep Transformer, an ablation study is conducted. Two ablative baselines are set up with partial components upon trained MST model: 1) \textbf{Ego:} MST that removes output level Transformer for the ego vehicle's route waypoints $D^{ego}_{Kt}$; 2) \textbf{Neighbors (N):} MST that removes all components related to $\mathcal{K}_t$; 3) \textbf{Neighbor-SLT (N-SLT)}: Retrained baseline of neighbors boosted by SLT. All ablative baselines are tested with identical testing setups in Section \ref{testing_sec}, and the results are displayed in Table \ref{table_5}. 

\begin{table}[tb]
\caption{Ablative Studies for Proposed Framework}
\centering
\setlength{\tabcolsep}{1.2mm}{

\begin{tabular}{c|c|cccccc}
\toprule
\multicolumn{2}{c}{}                      & Proposed & MST & Ego & N-SLT& N & SAC \\\midrule
\multirow{2}{*}{\textbf{Left Turn}}      & Succ.\% & 94       & 88   & 78 & 68         & 56             & 68  \\
                                & Coll.\% & 4        & 12   & 20          & 24 & 36             & 28  \\\midrule
\multirow{2}{*}{\textbf{Double Merging}} & Succ.\% & 96       & 92   & 90 &  80       & 72             & 62  \\
                                & Coll.\% & 2        & 4    & 4           & 20 & 25             & 22  \\\midrule
\multirow{2}{*}{\textbf{Roundabout-A}}    & Succ.\% & 88       & 84   & 82  & 74       & 68             & 76  \\
                                & Coll.\% & 12       & 16   & 14       & 26 & 32             & 24  \\\midrule
\multirow{2}{*}{\textbf{Roundabout-B}}    & Succ.\% & 82       & 76   & 72 & 70          & 44             & 48  \\
                                & Coll.\% & 18       & 24   & 28      & 30   & 56             & 52  \\\midrule
\multirow{2}{*}{\textbf{Roundabout-C}}    & Succ.\% & 76       & 66    & 62  & 66       & 34             & 46  \\
                                & Coll.\% & 24       & 34   & 38       & 32  & 64             & 48   \\\bottomrule 
\end{tabular}

}

\label{table_5}
\end{table}

The results of the Ego baseline imply that candidate route waypoints $\mathcal{K}^{ego}_t$ is the key for performance improvement of MST compared with efficient RL baseline such as DrQ. This is because the ego agent's route waypoints offer a filtered area for the ego vehicle to present the driving intention given information of neighbor vehicles and their corresponding intentions ($\mathcal{K}^{n}_t$), enhancing the scene understanding ability. Therefore, there is a significant performance drop without $\mathcal{K}^{ego}_t$ in these scenarios that have multiple potential intentions (Left Turn). Moreover, the poor results from the Neighbors baseline indicate that scene understanding is more dependent on $\mathcal{K}^{n}_t$ in the case of MST, as the local intentions for neighboring vehicles are more informative to the ego agent's decision-making. The results of attention weights in Fig. \ref{fig12} indicate that a multi-stage Transformer-based structure is the key to having the capability to understand intentions and interactions in urban scenes with adaptive information flow by attention. 

The role of global predictive representation learning lies in its guidance for lessening the state-action exploration space $\mathcal{S}\times\mathcal{A}$. Trained by an auxiliary representation learning objective through SLT, the latent representations $h_t$ for decision-making would share the sequential consistent features conditioned on actions. It can discover the overall picture of Q-value and lessen state-action space, lifting the sample efficiency of policy $\pi$ simultaneously. Therefore, the improvement of driving policy is reflected in 1) better quantitative performance (Table \ref{table_4}, Table \ref{table_5}); and 2) more diversified driving maneuvers to reach the destination. The former is counted on the fact that the global prediction guides the ego vehicle's policy to discover the advantageous regions of the Q-value more efficiently. The latter is inferred by the fact that there are possibly multiple optimal trajectories for driving maneuvers to reach the destination within close time steps.

\begin{table}[tbp]
\vspace{-0.2cm}
\caption{Ablative evaluations for SLT and state representations}
\centering
\setlength{\tabcolsep}{2.5mm}{
\begin{tabular}{l|l|lll}
\toprule
\textbf{Methods}               & \textbf{Scenario} & $\textbf{Norm} L_2$ & $\textbf{maxQ}$ & $\textbf{minQ}$ \\\midrule
\multirow{2}{*}{\textbf{Proposed}}      & left turn         & 0.756            & 1.386          & 0.857          \\
                               & merge             & 0.764            & 1.212          & 0.787          \\\midrule
\multirow{2}{*}{MST}           & left turn         & 0.682            & 1.235          & 0.411          \\
                               & merge             & 0.619            & 1.18           & 0.748          \\\midrule
\multirow{2}{*}{CNN+SLT}       & left turn         & 0.661            & 1.137          & -1.361         \\
                               & merge             & 0.47             & 1.174          & -0.975         \\\midrule
\multirow{2}{*}{CNN}           & left turn         & 0.522            & 0.742          & -1.31          \\
                               & merge             & 0.427            & 1.135          & -1.012         \\\midrule
\multirow{2}{*}{Neighbors+SLT} & left turn         & 0.492            & 0.971          & 0.246          \\
                               & merge             & 0.612            & 1.198          & -0.109         \\\midrule
\multirow{2}{*}{Neighbors}     & left turn         & 0.422            & 0.668          & 0.236          \\
                               & merge             & 0.175            & 1.193          & 0.975          \\\bottomrule
\end{tabular}}
\vspace{-0.1cm}
\label{table-vi}
\end{table}

\begin{figure*}[tp]
    \centering
    \includegraphics[width=\linewidth]{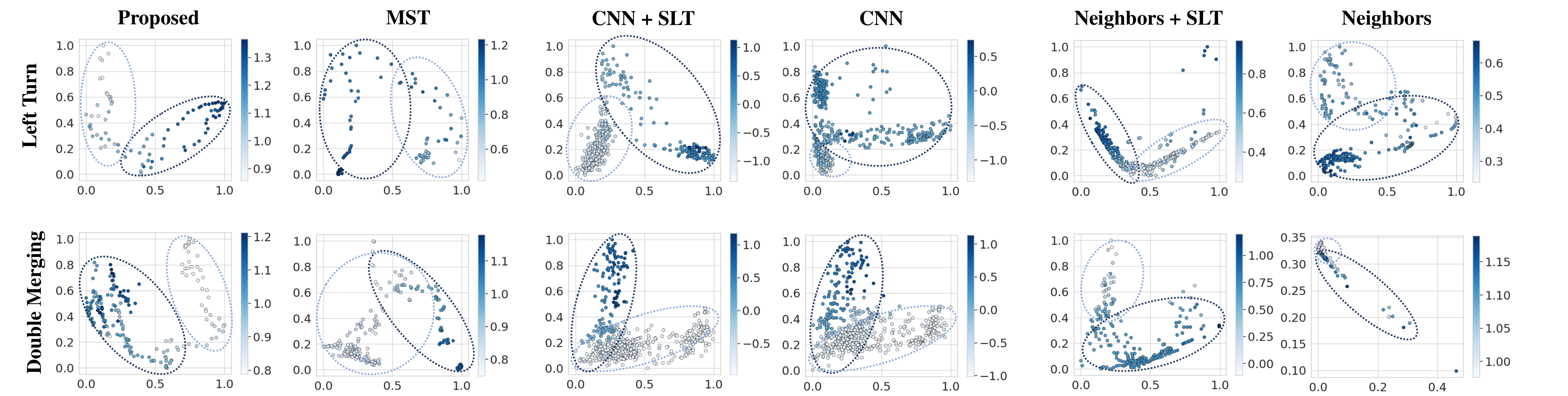}
    \caption{Principle Component Analysis (PCA) of the first two dimensions of the learned latent representation $\mathcal{S}\times\mathcal{A}$ by Scene-Rep Transformer and MST only. The color bar for each point is mapped with the corresponding mean Q-value by double Q-networks.}
    \vspace{-0.5cm}
    \label{fig:fig11}
\end{figure*}

To further justify the former claim of Q-value efficiency, the Q-value after global prediction learning (SLT) should be more separable from plain MST in the exploration space. Therefore, a PCA is conducted given the mean concatenate inputs for double Q-networks representing all possible explorations in $\mathcal{S} \times \mathcal{A}$ and the mean of the matching Q-values. The normalized first two components of PCA are represented for the concatenated input by state and action. To quantify the separability, we measure the $L_2$ distance between the Q-value groups ($25th-75th$ percentile) and the extremum. To further examine the mechanism of SLT, here we compare extra state representations of 1) CNN: rasterized BEV images encoded through CNNs; 2) Neighbor: vectorized trajectories of the ego vehicle and neighbors. Table \ref{table-vi} clearly reflects an overall improvement of separability ($10.8\%-26.6\%$), as well as the lifting of Q-values after adding global prediction learning (SLT) to each state representation. Qualitative results in Fig. \ref{fig:fig11} show a better-separating capability of the proposed method 
\begin{figure}[tp]
    \centering
    \includegraphics[width=\linewidth]{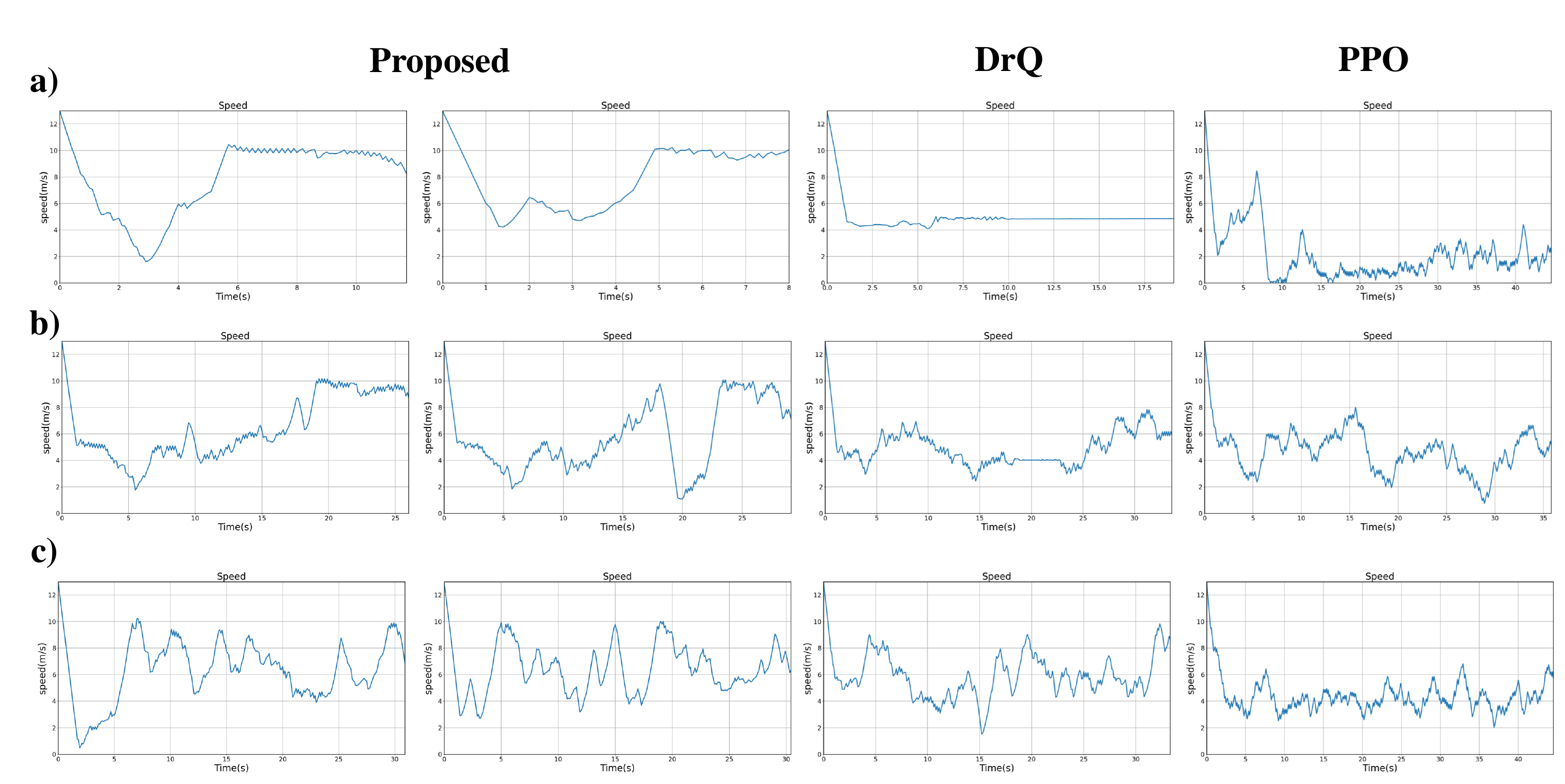}
    \caption{Principle speed patterns of ego vehicle for successful testing trials by Scene-Rep Transformer, DrQ, and PPO baseline. Three methods are tested in a) Unprotected left turn; b) Double Merging; c) Roundabout-B. }
    \label{fig:fig10}
\end{figure}
\begin{figure}[tp]
    \centering
    \includegraphics[width=\linewidth]{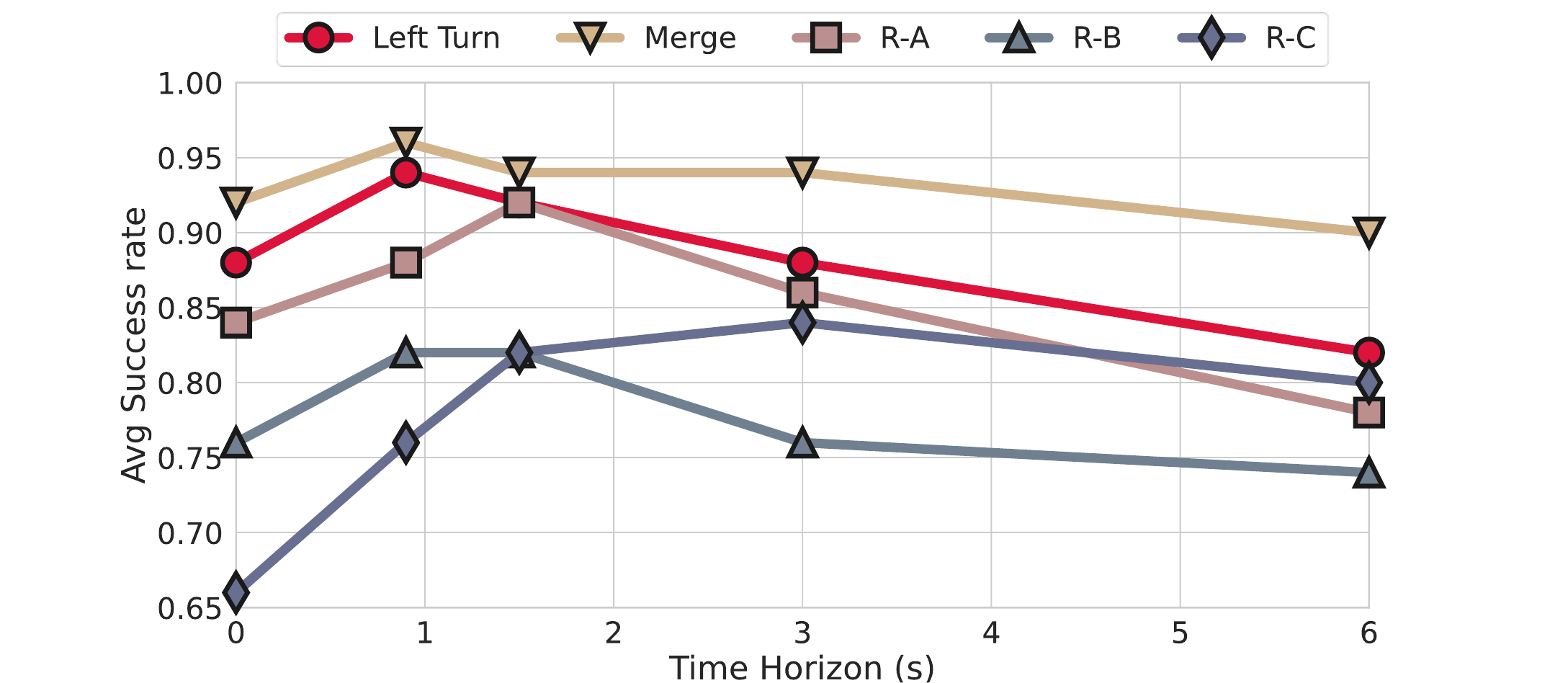}
    \caption{Comparison of selecting different $T_G$ for SLT training. Peak performance comes later given longer type of driving tasks.}
    \label{fig:fig-a}
\end{figure}

To manifest the latter claim for global predictive knowledge, trajectories of successful trials during testing are collected and shown in Fig. \ref{fig:fig.9}. The driving trajectories in each urban scenario corroborate the claim of the diversity of driving maneuvers. For instance, the ego vehicle in an unprotected left turn learns not only a "shortcut" to turning left to the inner lane and executes lane changing, but also a direct left turn to the second lane. More frequent lane changing occurs in different positions of the task for double merging and roundabout. Notably, the driving trajectories are more aligned with the center of each lane compared with DrQ which also displays somewhat diversity in certain urban scenarios. This characteristic might be owing to the center waypoint information by local predictions. For PPO, the previous claim for overfitted maneuvers is also verified by the monotonous trajectory distribution of PPO.

\begin{figure}[tp]
    \centering
    \includegraphics[width=\linewidth]{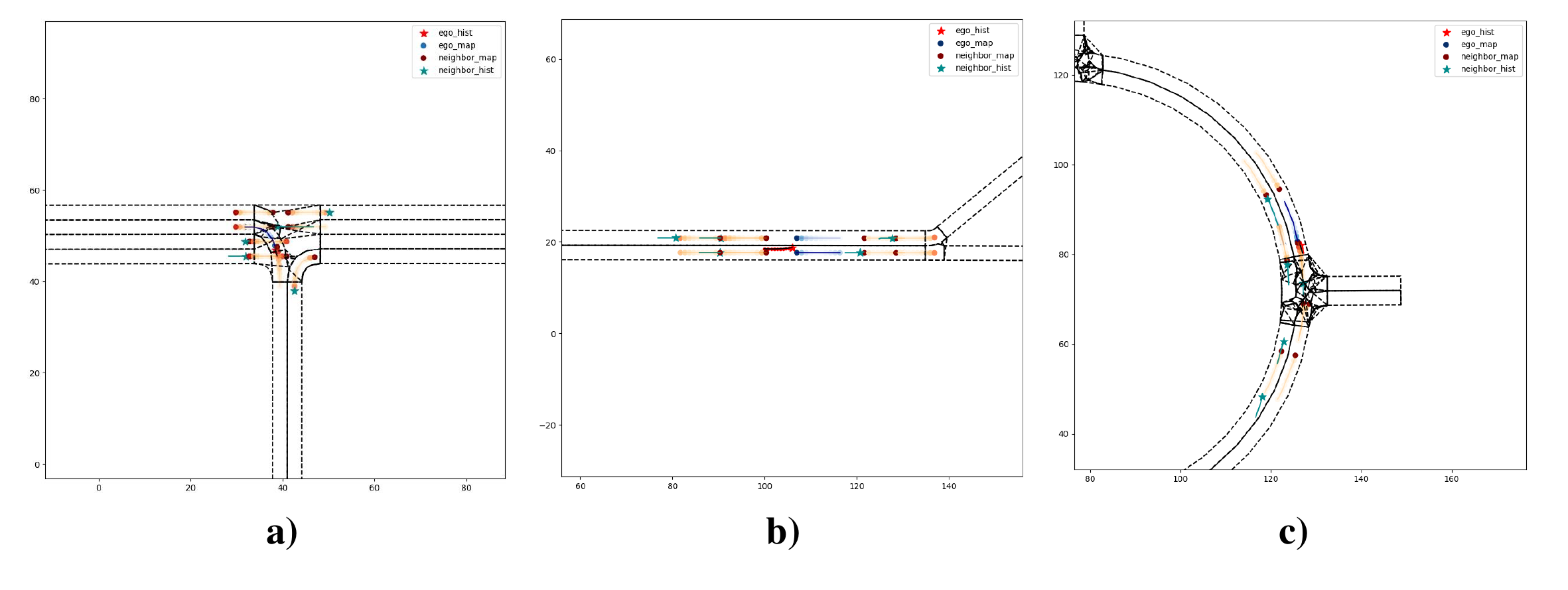}
    \vspace{-0.4cm}
    \caption{Attention visualization results in a) unprotected left turn, b)double merging, c) roundabout. The information with intentions by candidate waypoints and interactions are adjusted with adaptive information flows through the multi-stage Transformer.}
    \label{fig12}
\end{figure}

To investigate the proper future horizon settings for different lengths of tasks, in Fig.\ref{fig:fig-a} we compare the testing success rate of each scenario trained with SLT of different future horizons $T_G=(0,3,5,10,20)$. Results indicate faster improvements given a short future horizon and converges if $T_G$ is too large. The peak of performance comes later for a longer length of the task given (R-C). It is due to extra verbosity and demands for frequent updates of the predictive learning process. For long-term tasks, choosing a longer $T_G$ results in better performance compared to Table \ref{table_4}.

The diversity of driving maneuvers is also reflected by the bountiful patterns of vehicle dynamic states displayed in Fig. \ref{fig:fig10}. At least two significant patterns can be seen for the Scene-Rep Transformer. For instance, in an unprotected left turn (Fig. \ref{fig:fig10}a), the ego vehicle has learned the conservative dynamic patterns to nudge with a low speed in the intersection, change the lane, and remain at a constant high speed after changing to the second lane. It has also learned an aggressive manner without lane change. For the merging cases, the ego vehicle is capable of frequent lane changing to reach the destination, or just conducting a single lane change and waiting for the front vehicle but reaching the destination at similar time steps. The performances of DrQ and PPO baselines are alike and incompetent, as they cannot effectively reduce the exploration space. 


\section{Conclusions and Future Work}
This paper proposes the Scene-Rep Transformer, a novel Transformer-based representation learning framework that enhances the sample efficiency and performance of RL decision-making. The framework consists of two main blocks: a multi-stage Transformer for encoding the multi-modal scene state input and acquiring a representation understanding of the interactive driving scenes, and a sequential latent Transformer for distilling sequential predictive information into the current latent vector to guide the decision-making process. A soft actor-critic (SAC) decision-making module takes as input the refined latent representation and generates decision-making outputs. All the modules are integrated and trained end-to-end with the aim of obtaining a higher reward and success rate. We conduct extensive validation with five challenging simulated urban driving scenarios. Both quantitative and qualitative results reflect the overall superior performance of the proposed method in terms of sample efficiency, testing performance, interpretability, and diversity of driving maneuvers. We also investigate the effects of different components in the framework and find that the multi-stage Transformer is adaptively aware of intentions and interactions between agents and maps, and the sequential latent Transformer can effectively reduce the exploration space.

Future work will focus on improving or dynamically adjusting the depth of global latent predictions, so as to reduce the noise induced by longer-term predictions. We also aim to explicitly consider prediction uncertainties and incorporate our framework into an end-to-end driving system.

\bibliographystyle{IEEEtran}
\bibliography{b1}

\begin{thebibliography}{10}
\providecommand{\url}[1]{#1}
\csname url@samestyle\endcsname
\providecommand{\newblock}{\relax}
\providecommand{\bibinfo}[2]{#2}
\providecommand{\BIBentrySTDinterwordspacing}{\spaceskip=0pt\relax}
\providecommand{\BIBentryALTinterwordstretchfactor}{4}
\providecommand{\BIBentryALTinterwordspacing}{\spaceskip=\fontdimen2\font plus
\BIBentryALTinterwordstretchfactor\fontdimen3\font minus
  \fontdimen4\font\relax}
\providecommand{\BIBforeignlanguage}[2]{{%
\expandafter\ifx\csname l@#1\endcsname\relax
\typeout{** WARNING: IEEEtran.bst: No hyphenation pattern has been}%
\typeout{** loaded for the language `#1'. Using the pattern for}%
\typeout{** the default language instead.}%
\else
\language=\csname l@#1\endcsname
\fi
#2}}
\providecommand{\BIBdecl}{\relax}
\BIBdecl

\bibitem{huang2021driving}
Z.~Huang, J.~Wu, and C.~Lv, ``Driving behavior modeling using naturalistic
  human driving data with inverse reinforcement learning,'' \emph{IEEE
  Transactions on Intelligent Transportation Systems}, 2021.

\bibitem{b2}
L.~Zhu, F.~R. Yu, Y.~Wang, B.~Ning, and T.~Tang, ``Big data analytics in
  intelligent transportation systems: A survey,'' \emph{IEEE Transactions on
  Intelligent Transportation Systems}, vol.~20, no.~1, pp. 383--398, 2018.

\bibitem{b3}
E.~Yurtsever, J.~Lambert, A.~Carballo, and K.~Takeda, ``A survey of autonomous
  driving: Common practices and emerging technologies,'' \emph{IEEE access},
  vol.~8, pp. 58\,443--58\,469, 2020.

\bibitem{b4}
B.~R. Kiran, I.~Sobh, V.~Talpaert, P.~Mannion, A.~A. Al~Sallab, S.~Yogamani,
  and P.~P{\'e}rez, ``Deep reinforcement learning for autonomous driving: A
  survey,'' \emph{IEEE Transactions on Intelligent Transportation Systems},
  2021.

\bibitem{huang2022efficient}
Z.~Huang, J.~Wu, and C.~Lv, ``Efficient deep reinforcement learning with
  imitative expert priors for autonomous driving,'' \emph{IEEE Transactions on
  Neural Networks and Learning Systems}, 2022.

\bibitem{wu2021prioritized}
J.~Wu, Z.~Huang, W.~Huang, and C.~Lv, ``Prioritized experience-based
  reinforcement learning with human guidance: Methdology and application to
  autonomous driving,'' \emph{arXiv preprint arXiv:2109.12516}, 2021.

\bibitem{b5}
S.~Aradi, ``Survey of deep reinforcement learning for motion planning of
  autonomous vehicles,'' \emph{IEEE Transactions on Intelligent Transportation
  Systems}, 2020.

\bibitem{b6}
J.~Chen, B.~Yuan, and M.~Tomizuka, ``Model-free deep reinforcement learning for
  urban autonomous driving,'' in \emph{2019 IEEE intelligent transportation
  systems conference (ITSC)}.\hskip 1em plus 0.5em minus 0.4em\relax IEEE,
  2019, pp. 2765--2771.

\bibitem{huang2020multi}
Z.~Huang, C.~Lv, Y.~Xing, and J.~Wu, ``Multi-modal sensor fusion-based deep
  neural network for end-to-end autonomous driving with scene understanding,''
  \emph{IEEE Sensors Journal}, vol.~21, no.~10, pp. 11\,781--11\,790, 2020.

\bibitem{chen2019deep}
J.~Chen, B.~Yuan, and M.~Tomizuka, ``Deep imitation learning for autonomous
  driving in generic urban scenarios with enhanced safety,'' in \emph{2019
  IEEE/RSJ International Conference on Intelligent Robots and Systems
  (IROS)}.\hskip 1em plus 0.5em minus 0.4em\relax IEEE, 2019, pp. 2884--2890.

\bibitem{wu2021human}
J.~Wu, Z.~Huang, C.~Huang, Z.~Hu, P.~Hang, Y.~Xing, and C.~Lv,
  ``Human-in-the-loop deep reinforcement learning with application to
  autonomous driving,'' \emph{arXiv preprint arXiv:2104.07246}, 2021.

\bibitem{b12}
H.~Liu, Z.~Huang, and C.~Lv, ``Improved deep reinforcement learning with expert
  demonstrations for urban autonomous driving,'' \emph{arXiv preprint
  arXiv:2102.09243}, 2021.

\bibitem{perez2022deep}
{\'O}.~P{\'e}rez-Gil, R.~Barea, E.~L{\'o}pez-Guill{\'e}n, L.~M. Bergasa,
  C.~Gomez-Huelamo, R.~Guti{\'e}rrez, and A.~Diaz-Diaz, ``Deep reinforcement
  learning based control for autonomous vehicles in carla,'' \emph{Multimedia
  Tools and Applications}, vol.~81, no.~3, pp. 3553--3576, 2022.

\bibitem{wolf2017learning}
P.~Wolf, C.~Hubschneider, M.~Weber, A.~Bauer, J.~H{\"a}rtl, F.~D{\"u}rr, and
  J.~M. Z{\"o}llner, ``Learning how to drive in a real world simulation with
  deep q-networks,'' in \emph{2017 IEEE Intelligent Vehicles Symposium
  (IV)}.\hskip 1em plus 0.5em minus 0.4em\relax IEEE, 2017, pp. 244--250.

\bibitem{feher2019hybrid}
{\'A}.~Feh{\'e}r, S.~Aradi, F.~Heged{\"u}s, T.~B{\'e}csi, and
  P.~G{\'a}sp{\'a}r, ``Hybrid ddpg approach for vehicle motion planning,''
  2019.

\bibitem{duan2020hierarchical}
J.~Duan, S.~E. Li, Y.~Guan, Q.~Sun, and B.~Cheng, ``Hierarchical reinforcement
  learning for self-driving decision-making without reliance on labelled
  driving data,'' \emph{IET Intelligent Transport Systems}, vol.~14, no.~5, pp.
  297--305, 2020.

\bibitem{janner2021reinforcement}
M.~Janner, Q.~Li, and S.~Levine, ``Reinforcement learning as one big sequence
  modeling problem,'' in \emph{ICML 2021 Workshop on Unsupervised Reinforcement
  Learning}, 2021.

\bibitem{chen2021decision}
L.~Chen, K.~Lu, A.~Rajeswaran, K.~Lee, A.~Grover, M.~Laskin, P.~Abbeel,
  A.~Srinivas, and I.~Mordatch, ``Decision transformer: Reinforcement learning
  via sequence modeling,'' \emph{Advances in neural information processing
  systems}, vol.~34, pp. 15\,084--15\,097, 2021.

\bibitem{bojarski2016end}
M.~Bojarski, D.~Del~Testa, D.~Dworakowski, B.~Firner, B.~Flepp, P.~Goyal, L.~D.
  Jackel, M.~Monfort, U.~Muller, J.~Zhang \emph{et~al.}, ``End to end learning
  for self-driving cars,'' \emph{arXiv preprint arXiv:1604.07316}, 2016.

\bibitem{dosovitskiy2017carla}
A.~Dosovitskiy, G.~Ros, F.~Codevilla, A.~Lopez, and V.~Koltun, ``Carla: An open
  urban driving simulator,'' in \emph{Conference on robot learning}.\hskip 1em
  plus 0.5em minus 0.4em\relax PMLR, 2017, pp. 1--16.

\bibitem{casas2021mp3}
S.~Casas, A.~Sadat, and R.~Urtasun, ``Mp3: A unified model to map, perceive,
  predict and plan,'' in \emph{Proceedings of the IEEE/CVF Conference on
  Computer Vision and Pattern Recognition}, 2021, pp. 14\,403--14\,412.

\bibitem{b14}
Z.~Huang, X.~Mo, and C.~Lv, ``Multi-modal motion prediction with
  transformer-based neural network for autonomous driving,'' \emph{arXiv
  preprint arXiv:2109.06446}, 2021.

\bibitem{b13}
J.~Gao, C.~Sun, H.~Zhao, Y.~Shen, D.~Anguelov, C.~Li, and C.~Schmid,
  ``Vectornet: Encoding hd maps and agent dynamics from vectorized
  representation,'' in \emph{Proceedings of the IEEE/CVF Conference on Computer
  Vision and Pattern Recognition}, 2020, pp. 11\,525--11\,533.

\bibitem{b15}
J.~Ngiam, B.~Caine, V.~Vasudevan, Z.~Zhang, H.-T.~L. Chiang, J.~Ling,
  R.~Roelofs, A.~Bewley, C.~Liu, A.~Venugopal \emph{et~al.}, ``Scene
  transformer: A unified multi-task model for behavior prediction and
  planning,'' \emph{arXiv e-prints}, pp. arXiv--2106, 2021.

\bibitem{gilles2022gohome}
T.~Gilles, S.~Sabatini, D.~Tsishkou, B.~Stanciulescu, and F.~Moutarde,
  ``Gohome: Graph-oriented heatmap output for future motion estimation,'' in
  \emph{2022 International Conference on Robotics and Automation (ICRA)}.\hskip
  1em plus 0.5em minus 0.4em\relax IEEE, 2022, pp. 9107--9114.

\bibitem{bender2014lanelets}
P.~Bender, J.~Ziegler, and C.~Stiller, ``Lanelets: Efficient map representation
  for autonomous driving,'' in \emph{2014 IEEE Intelligent Vehicles Symposium
  Proceedings}.\hskip 1em plus 0.5em minus 0.4em\relax IEEE, 2014, pp.
  420--425.

\bibitem{b18-2}
M.~H{\"u}gle, G.~Kalweit, M.~Werling, and J.~Boedecker, ``Dynamic
  interaction-aware scene understanding for reinforcement learning in
  autonomous driving,'' in \emph{2020 IEEE International Conference on Robotics
  and Automation (ICRA)}.\hskip 1em plus 0.5em minus 0.4em\relax IEEE, 2020,
  pp. 4329--4335.

\bibitem{b18-3}
P.~Cai, H.~Wang, Y.~Sun, and M.~Liu, ``Dq-gat: Towards safe and efficient
  autonomous driving with deep q-learning and graph attention networks,''
  \emph{arXiv preprint arXiv:2108.05030}, 2021.

\bibitem{b17}
Y.~Liu, J.~Zhang, L.~Fang, Q.~Jiang, and B.~Zhou, ``Multimodal motion
  prediction with stacked transformers,'' in \emph{Proceedings of the IEEE/CVF
  Conference on Computer Vision and Pattern Recognition}, 2021, pp. 7577--7586.

\bibitem{b23}
D.~Graves, N.~M. Nguyen, K.~Hassanzadeh, and J.~Jin, ``Learning predictive
  representations in autonomous driving to improve deep reinforcement
  learning,'' \emph{arXiv preprint arXiv:2006.15110}, 2020.

\bibitem{b27}
Y.~Xiao, F.~Codevilla, C.~Pal, and A.~M. L{\'o}pez, ``Action-based
  representation learning for autonomous driving,'' \emph{arXiv preprint
  arXiv:2008.09417}, 2020.

\bibitem{b28}
J.-B. Grill, F.~Strub, F.~Altch{\'e}, C.~Tallec, P.~Richemond, E.~Buchatskaya,
  C.~Doersch, B.~Avila~Pires, Z.~Guo, M.~Gheshlaghi~Azar \emph{et~al.},
  ``Bootstrap your own latent-a new approach to self-supervised learning,''
  \emph{Advances in Neural Information Processing Systems}, vol.~33, pp.
  21\,271--21\,284, 2020.

\bibitem{b29}
D.~Yarats, I.~Kostrikov, and R.~Fergus, ``Image augmentation is all you need:
  Regularizing deep reinforcement learning from pixels,'' in
  \emph{International Conference on Learning Representations}, 2020.

\bibitem{b30}
A.~Srinivas, M.~Laskin, and P.~Abbeel, ``Curl: Contrastive unsupervised
  representations for reinforcement learning,'' \emph{arXiv preprint
  arXiv:2004.04136}, 2020.

\bibitem{b25}
T.~Wang, Y.~Luo, J.~Liu, R.~Chen, and K.~Li, ``End-to-end self-driving approach
  independent of irrelevant roadside objects with auto-encoder,'' \emph{IEEE
  Transactions on Intelligent Transportation Systems}, vol.~23, no.~1, pp.
  641--650, 2020.

\bibitem{b26}
K.~Chen, L.~Hong, H.~Xu, Z.~Li, and D.-Y. Yeung, ``Multisiam: Self-supervised
  multi-instance siamese representation learning for autonomous driving,'' in
  \emph{Proceedings of the IEEE/CVF International Conference on Computer
  Vision}, 2021, pp. 7546--7554.

\bibitem{b31}
M.~Schwarzer, A.~Anand, R.~Goel, R.~D. Hjelm, A.~Courville, and P.~Bachman,
  ``Data-efficient reinforcement learning with self-predictive
  representations,'' \emph{arXiv preprint arXiv:2007.05929}, 2020.

\bibitem{b33}
T.~Haarnoja, A.~Zhou, K.~Hartikainen, G.~Tucker, S.~Ha, J.~Tan, V.~Kumar,
  H.~Zhu, A.~Gupta, P.~Abbeel \emph{et~al.}, ``Soft actor-critic algorithms and
  applications,'' \emph{arXiv preprint arXiv:1812.05905}, 2018.

\bibitem{b34}
X.~Chen and K.~He, ``Exploring simple siamese representation learning,'' in
  \emph{Proceedings of the IEEE/CVF Conference on Computer Vision and Pattern
  Recognition}, 2021, pp. 15\,750--15\,758.

\bibitem{b35}
C.~Gelada, S.~Kumar, J.~Buckman, O.~Nachum, and M.~G. Bellemare, ``Deepmdp:
  Learning continuous latent space models for representation learning,'' in
  \emph{International Conference on Machine Learning}.\hskip 1em plus 0.5em
  minus 0.4em\relax PMLR, 2019, pp. 2170--2179.

\bibitem{alemi2016deep}
A.~A. Alemi, I.~Fischer, J.~V. Dillon, and K.~Murphy, ``Deep variational
  information bottleneck,'' \emph{arXiv preprint arXiv:1612.00410}, 2016.

\bibitem{b32}
M.~Zhou, J.~Luo, J.~Villella, Y.~Yang, D.~Rusu, J.~Miao, W.~Zhang, M.~Alban,
  I.~Fadakar, Z.~Chen \emph{et~al.}, ``Smarts: Scalable multi-agent
  reinforcement learning training school for autonomous driving,'' \emph{arXiv
  preprint arXiv:2010.09776}, 2020.

\bibitem{wang2020learning}
J.~Wang, Y.~Wang, D.~Zhang, Y.~Yang, and R.~Xiong, ``Learning hierarchical
  behavior and motion planning for autonomous driving,'' in \emph{2020 IEEE/RSJ
  International Conference on Intelligent Robots and Systems (IROS)}.\hskip 1em
  plus 0.5em minus 0.4em\relax IEEE, 2020, pp. 2235--2242.

\bibitem{b36}
J.~Schulman, F.~Wolski, P.~Dhariwal, A.~Radford, and O.~Klimov, ``Proximal
  policy optimization algorithms,'' \emph{arXiv preprint arXiv:1707.06347},
  2017.

\bibitem{b10}
D.~Zhu, T.~Li, D.~Ho, C.~Wang, and M.~Q.-H. Meng, ``Deep reinforcement learning
  supervised autonomous exploration in office environments,'' in \emph{2018
  IEEE international conference on robotics and automation (ICRA)}.\hskip 1em
  plus 0.5em minus 0.4em\relax IEEE, 2018, pp. 7548--7555.

\bibitem{tishby2000information}
N.~Tishby, F.~C. Pereira, and W.~Bialek, ``The information bottleneck method,''
  \emph{arXiv preprint physics/0004057}, 2000.

\bibitem{nakamura2022representation}
H.~Nakamura, M.~Okada, and T.~Taniguchi, ``Representation uncertainty in
  self-supervised learning as variational inference,'' \emph{arXiv preprint
  arXiv:2203.11437}, 2022.

\bibitem{hafner2019dream}
D.~Hafner, T.~Lillicrap, J.~Ba, and M.~Norouzi, ``Dream to control: Learning
  behaviors by latent imagination,'' \emph{arXiv preprint arXiv:1912.01603},
  2019.

\bibitem{poole2019variational}
B.~Poole, S.~Ozair, A.~Van Den~Oord, A.~Alemi, and G.~Tucker, ``On variational
  bounds of mutual information,'' in \emph{International Conference on Machine
  Learning}.\hskip 1em plus 0.5em minus 0.4em\relax PMLR, 2019, pp. 5171--5180.

\bibitem{okada2021dreaming}
M.~Okada and T.~Taniguchi, ``Dreaming: Model-based reinforcement learning by
  latent imagination without reconstruction,'' in \emph{2021 ieee international
  conference on robotics and automation (icra)}.\hskip 1em plus 0.5em minus
  0.4em\relax IEEE, 2021, pp. 4209--4215.

\end{thebibliography}

\appendix
\section{\Ch{Appendix}}
\label{appendix}
\begin{theorem}[SLT's equivalence to information bottlenecks]
    \label{theorem}
     Proposed SLT pipeline with Sim-Siam \cite{b10} representation learning objectives equals to the future state information bottleneck objectives for latent dynamic models \cite{tishby2000information}:
        \begin{equation}
        \label{eib}
        J:\max \left[ I(h_{1:T},s_{1:T}|a_{1:T}) - \beta I(h_{1:T},s^i_{1:T}|a_{1:T}) \right]
        \end{equation}
    Where $T=T_f$ stands for the future horizons. $s\in\mathcal{S}$, $a\in\mathcal{A}$, and $h\in\ D$ denotes states, actions and representations for driving scene. $i$ is the batch sampled index \cite{alemi2016deep}.
\end{theorem}

\begin{proof}
    According to Lemma \ref{lemma1}, we can derive the objective lower bound: 
        $J \geq \sum_t \left(\mathcal{L}^{NCE}_t + \beta \mathcal{L}^{KL}_t \right) $.
    From Lemma \ref{lemma3}, we proof the validity of $\mathcal{L}^{NCE}_t$ through SLT. Then, as the equivalence of $\mathcal{L}^{NCE}_t$ with Sim-Siam pipeline proved in Lemma \ref{lemma2}, and $\mathcal{L}^{KL}_t=\operatorname{Const.}$ \cite{nakamura2022representation} as $h$ is deterministic in our representation learning pipelines, the theorem is proved.
\end{proof}

\begin{lemma}[Lower bound for information bottlenecks]
    \label{lemma1}
    The proposed objectives of information bottlenecks across future timesteps can be bounded by \cite{hafner2019dream}:
    \begin{equation}
        \label{elb}
        J \geq \sum_t \left(\mathcal{L}^{NCE}_t + \beta \mathcal{L}^{KL}_t \right)
    \end{equation}
    Where $\mathcal{L}^{NCE}_t$ denotes the batched InfoNCE objectives \cite{poole2019variational}:
    \begin{equation}
        \label{Lnce}
        \mathcal{L}_t^{NCE}=\underset{h\sim \Phi}{\mathbb{E}} \left[\log p\left(h_t \mid s_t\right)-\log \sum_{i} p\left(h_t \mid s^i_t\right)\right]
    \end{equation}
    $\Phi$ is proposed MST, and $\mathcal{L}^{KL}_t$ is the KL-divergence:
    \begin{equation}
        \operatorname{KL}\left(q\left(h_t \mid h_{t-1}, a_{t-1}, s_t\right) \| p\left(h_t \mid h_{t-1}, a_{t-1}\right)\right)
    \end{equation}
\end{lemma}
\begin{proof}
    Consider the first term of information gain:
     \begin{equation}
        \label{elb0}
        \begin{aligned}
        &I\left(h_{1: T} ; s_{1: T} \mid a_{1: T}\right) \\
        &=\mathbb{E}\left[\sum_t \log p\left(s_{1: T} \mid h_{1: T}, a_{1: T}\right)-
        \log p\left(s_{1: T}, \mid a_{1: T}\right)\right] \\
        &\geq \mathbb{E}\left[\sum_t \log p\left(s_{1: T} \mid h_{1: T}, a_{1: T}\right)\right]-\\
        &\qquad \mathrm{KL}\left(p\left(s_{1: T} \mid h_{1: T}, a_{1: T}\right) \| \prod_t p\left(s_t \mid h_t\right)\right) \\
        &=\underset{h\sim \Phi}{\mathbb{E}} \sum_t \log p\left(s_t \mid h_t\right)
        \end{aligned}
    \end{equation}
    The final term is derived due to the independence as $\log p(s_{1:T}|a_{1:T})=0$. By using Bayes rules and batched InfoNCE, the expected term for each future step becomes:
    \begin{equation}
        \label{infonce}
        \underset{h\sim \Phi}{\mathbb{E}}  \log p\left(s_t \mid h_t\right) \geq \mathcal{L}_t^{NCE}
    \end{equation}
    Consider the second term of information gain:
    \begin{equation}
        \begin{aligned}
        & I\left(h_{1: T} ; s_{1: T}^i \mid a_{1: T}\right) \\
        &=\mathbb{E}\left[\sum_t \log p\left(h_t \mid h_{t-1}, a_{t-1}, s_t\right)-\log p\left(h_t \mid h_{t-1}, a_{t-1}\right)\right] \\
        & \leq \mathbb{E}\left[\sum_t \log q\left(h_t \mid h_{t-1}, a_{t-1}, s_t\right)-\log p\left(h_t \mid h_{t-1}, a_{t-1}\right)\right] \\
        &=\sum_t \operatorname{KL}\left(q\left(h_t \mid h_{t-1}, a_{t-1}, s_t\right) \| p\left(h_t \mid h_{t-1}, a_{t-1}\right)\right)
        \end{aligned}
    \end{equation}
    Denoting the final term as $\sum_t\mathcal{L}_t^{KL}$, the proposed lower bound is proved.
\end{proof}

\begin{lemma}[Equivalence of InfoNCE to Sim-Siam] 
\label{lemma2}
Sim-siam representation learning pipeline maximizes $\mathcal{L}_t^{NCE}$ in Equ. \ref{Lnce} implicitly.
\end{lemma}
\begin{proof}
    Denote the terms in Equ. \ref{Lnce}:
    \begin{equation}
    \begin{aligned}
        &\underset{h\sim \Phi}{\mathbb{E}}\log p\left(h_t \mid s_t\right)=\mathcal{L}^1\\
         &\underset{h\sim \Phi}{\mathbb{E}}\log \sum_{i} p\left(h_t \mid s^i_t\right)=\mathcal{L}^2
    \end{aligned}
    \end{equation}
    From the theorem proved in \cite{nakamura2022representation}, we can conclude that $\mathcal{L}^1\simeq\mathcal{L}_{glb}$ in Equ.\ref{e14}, and the presence of $\mathcal{P}$ in Sim-Siam implicitly minimize $\mathcal{L}^2$. Hence, the lemma is proved.
\end{proof}

\begin{lemma}[Validity of SLT to InfoNCE]
\label{lemma3}
It is valid to reformulate $h$ in Equ.\ref{e14} through SLT $\mathcal{T}$, as future representations $h_{1:T}$ cannot be directly sampled by $\Phi$ without knowing ground-truth future states.
\end{lemma}
\begin{proof}
    To reformulate $\mathcal{L}_t^{NCE}$ using SLT transition model $\mathcal{T}$, for each timestep $t\in[1, T_f]$, we can always obtain the multi-step predictions $h_t\leftarrow p(h_t|h_{<t},a_{<t})$ from SLT $\mathcal{T}$. Batched predictions are formulated as $\underset{h\sim \mathcal{T}}{\mathbb{E}}p(h_t|h_{<t},a_{<t})$. Then the representations $h_t$ of each future timestep can be assigned for $\mathcal{L}_t^{NCE}$ by SLT $\mathcal{T}$ with importance sampling methods proved in \cite{okada2021dreaming}. Therefore, the lemma is proved.
\end{proof}

\textbf{Waypoints generation}: $\mathcal{K}_t$ requires graph-structured High-fidelity (HD) Maps $G_{\textbf{HD}}$ \cite{bender2014lanelets} and current states $\textbf{M}^{ego}_0,\textbf{M}^{1:n}_0$. That information is pre-processed by the perception module in real cases. we can online generate waypoints for ego and neighbors by the following algorithm:
\begin{algorithm}[htb]
	\caption{Waypoints Generation}
	\begin{algorithmic}[1] 
	\Require HD Maps: $G_{\textbf{HD}}=(e_{ij},i,j\in N)$; States        $\textbf{M}^{ego}_0,\textbf{M}^{1:n}_0$
        \Require Distance threshold $d$; Length threshold $l$
	\For{$m$  \textbf{in} $\{ego,1:n\}$} 
            \State Nodes $\textbf{n}^{1,2,\cdots}\in N, \|(x_{\textbf{n}},y_{\textbf{n}})-(x_{\textbf{M}^m_0},y_{\textbf{M}^m_0})\|_2 < d$ 
            \For{$\textbf{n}^p$  \textbf{in} $\textbf{n}^{1,2,\cdot}$} 
                \State Search raw waypoints within length $l$:
                \State$\textbf{k}^{m,p}_t \leftarrow \operatorname{DepthFirstSearch}(\textbf{n}^p,G_{\textbf{HD}},l)$
                \State Remove intersected and filter forward waypoints:
                \State$\textbf{k}^{m,p}_t \leftarrow \operatorname{InterRemove(\textbf{k}^{m,p}_t)}$
                \State$\textbf{k}^{m,p}_t \leftarrow \operatorname{ForwardFilter(\textbf{k}^{m,p}_t, \textbf{M}^m_0)}$
            \EndFor
            \State Output $K^{m}_t = \{\textbf{k}^{m,1}_t, \textbf{k}^{m,2}_t, \cdots\}$
        \EndFor
        \State \textbf{return} $\mathcal{K}_t = \{K^{ego}_{t}, K^{1}_{t}, K^{2}_{t}, \cdots, K^{n}_{t}\}$
		
	\end{algorithmic}
	\label{Ag2}
\end{algorithm}

\begin{table}[ht]
\caption{Settings (Default) for the official SMARTS scenarios}
\label{table_1}
\centering
\begin{tabular}{l|l|l|l}
\toprule
Scenario Name& \textbf{Left Turn}&\textbf{Double Merge}&\textbf{Roundabout}\\\midrule
Max Timesteps & 400 & 400 &  400/600/800 \\\
Flows (/route)&20  &40  &20    \\\
Vehicles (/flow)&4  &6  &4    \\\midrule
Imperfection&\multicolumn{3}{c}{$\mathcal{N}(U[0.3,0.7],0.1)$}\\\
Impatience&\multicolumn{3}{c}{$U[0,1]$}\\\
Cooperative&\multicolumn{3}{c}{$U[0,1]$}\\\bottomrule
\end{tabular}
\end{table}

\begin{table}[htp]
\vspace{-0.2cm}
\caption{Parameters used in the experiment}
\label{table_2}
\centering
\begin{tabular}{@{}lllll@{}}
\toprule
Notation                & Meaning                               & Value         \\ \toprule
$n$                     & Max number of neighboring vehicles    & 5             \\
$V_{max}$               & Speed limit (m/s)                     & 10             \\
$T_h$                   & Historical horizon                    & 10             \\
$T_{K}$                 & Route waypoints horizon                  & 10             \\
$T_{G}$                 & Global predictive horizon                 & 3             \\
$N_{\textbf{k}}$        & Number of candidate route waypoints  & 2          \\\midrule
$\gamma$                & Discount rate                         & 0.99         \\
$\lambda$               & Polyak averaging weight               & 0.005        \\
$\alpha$                & Initial entropy weight                & 1             \\
$N_{init}$              & Warm-up random searching step         & 5000        \\
$N_{\text{buffer}}$     & Replay buffer capacity                & 20000         \\
$N_{\mathcal{B}}$       & Batch size                            & 32            \\
$N_{train}$             & Training steps                        & 100000        \\
\bottomrule
\end{tabular}
\vspace{-0.2cm}
\end{table}

\end{document}